\documentclass{article}
\usepackage{iclr2027_conference,times}

\usepackage{amsmath,amsfonts,bm}
\usepackage{sidecap}
\sidecaptionvpos{table}{c}

\def\eqref#1{equation~\ref{#1}}

\def\1{\bm{1}}

\DeclareMathAlphabet{\mathsfit}{\encodingdefault}{\sfdefault}{m}{sl}
\SetMathAlphabet{\mathsfit}{bold}{\encodingdefault}{\sfdefault}{bx}{n}

\usepackage{hyperref,url,amsmath,amssymb,booktabs,graphicx}
\hypersetup{hidelinks, pdftitle={State Space Attention: Learning Context-Dependent Memory Retention}, pdfauthor={Ege \c{C}akar, Cengiz Pehlevan}}
\newcommand{\sg}{\operatorname{sg}}
\newcommand{\softmaxop}{\operatorname{softmax}}
\newcommand{\softplusop}{\operatorname{softplus}}

\title{State Space Attention:\\Learning Context-Dependent Memory\\ Retention}
\author{Ege \c{C}akar \\
Department of Statistics\\
Harvard University\\
Cambridge, MA 02138, USA \\
\texttt{ecakar@college.harvard.edu}
\And
Cengiz Pehlevan \\
John A. Paulson School of\\
Engineering and Applied Sciences\\
Harvard University\\
\texttt{cpehlevan@seas.harvard.edu}}
\iclrfinalcopy

\begin{document}
\maketitle
\lhead{Preprint}

\begin{abstract}
Standard attention's key-value cache grows linearly with context length. Bounding it raises the question of what a sequence model should retain for long-context reasoning. We argue this choice should be learned, and introduce State Space Attention (SSA), which learns to keep a bounded set of past representations and reads them with standard attention. A learned scorer rescores every chunk against newly arriving context and the rest of memory. A fixed-context Transformer is then the special case that always keeps the most recent tokens. We train the scorer and reader jointly, either moving gradually from a differentiable soft weighting of memories to hard retention, or with hard retention from the start. To produce signal for the scorer, we generate supervised samples during training, a technique we call recovery. We compare SSA with recurrent models, bounded-memory attention methods, and full-attention Transformers across retrieval, variable tracking, multi-hop reasoning, and language modeling. On our retrieval and reasoning benchmarks, SSA matches or comes within a few points of full attention on most tasks, far outperforms a Transformer with the same readable budget on every long-context retrieval task we evaluate, and matches or outperforms one with about 4 times that budget, while using constant inference memory. The largest remaining gap to full attention at small language model scale is on RULER. SSA makes a fixed number of tokens useful over a longer context, and since it reads memory with standard attention, we propose that it can be stacked with KV cache eviction methods. We also pretrain and post-train 400M-parameter models and measure how SSA's training and inference costs scale with context length and model size. Together, these experiments examine the capabilities and limitations of learned memory retention for long-context sequence modeling.
\end{abstract}

\begin{center}
Code: \url{https://github.com/Ege-Cakar/State-Space-Attention}
\end{center}

\section{Introduction}

Long-context reasoning requires a model to use information that appeared far earlier in a sequence. Transformers do this by keeping key--value representations of every preceding token \citep{vaswani2017attention}, so their cache grows with the sequence. State Space Models (SSMs) and recurrent linear attention instead update a fixed-size state as tokens arrive \citep{gu2023mamba,dao2024mamba2,katharopoulos2020linear}, which makes inference memory independent of context length but makes precise recall difficult, because successive inputs interfere within the same representation \citep{jelassi2024repeat}.

We ask whether a model can keep the contextual reasoning of attention under the same constant-memory constraint. A bounded state compresses the context as a whole, but it need not compress the tokens it holds: a model can keep exact token representations, as long as it caps how many it keeps. Seen this way, a Transformer with context length $R$ already has a bounded state, managed by a first-in, first-out rule that keeps the most recent $R$ tokens. We instead let the model choose which $R$ tokens of the seen context to keep. Besides learning how to read its history, the model must then learn what to keep before it knows which questions it will need to answer.

We introduce State Space Attention (SSA), an architecture and training recipe for this approach. It replaces the growing attention cache with a fixed-size memory and a learned retention policy. As the sequence advances, a \textit{scorer}, a small Transformer, compares incoming information with stored content and decides what to discard, and the \textit{language backbone} (or \textit{reader}), also a Transformer, reads the retained history together with recent context. In our experiments, the model solves tasks over contexts up to $24\times$ its readable budget (the longest we tested). The scorer and the backbone are trained jointly, end-to-end and from scratch, which recent work lists as an important future direction \citep{yang2026indexmem}; \citet{he2025lte} showed a similar approach with a less expressive scorer.

SSA has no separate recurrent memory or external archive at inference: its persistent state is the retained representations, and its reader is standard attention over that state. We therefore propose that SSA composes with the methods we compare it with. Cache eviction and compression methods start from a model that reads the whole context and reduce what it stores; SSA takes a model with a fixed readable budget of $R$ tokens and extends the context over which those tokens are useful for retrieval and reasoning. An eviction or sparsification method can then be applied to SSA's retained set to shrink it further, and SSA layers can be stacked with recurrent layers (Table~\ref{tab:ssa_hybrids}); we test the latter at the 3M scale and leave the former untested. The retained keys and values of deeper layers were also computed with earlier context in view, so they carry lossy information about evicted content (\S\ref{sec:lossy}).

The main difficulty in training is credit assignment for the scorer. A useful memory is often discarded long before the prediction that reveals its value, and a hard selection provides no ordinary gradient. Straight-through estimation \citep{bengio2013ste} gives gradients for values that happen to be retained, but not for discarded ones, and a reader that has not learned to use memory cannot give a reliable retention signal. Our recipe develops the reader through language pretraining, then a recovery mechanism tests whether keeping a discarded chunk would have corrected a training error and credits the scorer when it would. We also propose an approximation to recovery: a soft phase in which the scorer down-weights chunks without removing them, followed by discrete retention. In our 400M ablations, no run without recovery learned retrieval, while runs without the soft phase did (\S\ref{sec:ablations}).

Our experiments test whether a bounded attention reader can solve a task given a perfect policy, whether a learned policy preserves useful information across a long sequence, and whether training reliably develops both. SSA comes within 1.3 points of the full-attention baseline on long-context cued retrieval and matches it on multi-hop reasoning with fresh data. Comparisons with recurrent models show that the memory representation matters: at similar state budgets, SSA stays at Transformer accuracy on multi-hop induction that fits its readable state when trained on fresh data, at 20M and at 400M, where GDN and HOLA do not. When the source of a multi-hop question must first survive eviction in a long input, SSA stays within one point of full attention in token accuracy if the source is at the start, where the other bounded models stay at chance; no model solves it with the source at a random position (\S\ref{sec:scaling_posttraining}).

Our contributions are a bounded attention architecture, a procedure for training its reader and retention policy jointly, an analysis of their capability and optimization trade-offs, and a comparison with leading constant-memory architectures. We establish task performance on controlled benchmarks at small scales, then pretrain 400M-parameter models on 15B tokens, post-train them on long-context tasks, and measure training and inference cost across context lengths and model sizes.

\section{Related work}
\label{sec:related}

\paragraph{Fixed-state sequence models.} Linear attention and state space models replace a growing context cache with an online state update \citep{katharopoulos2020linear,gu2023mamba,dao2024mamba2}. DeltaNet and Gated DeltaNet develop associative recurrent memories with learned updates \citep{yang2024deltanet,yang2025gateddeltanet}. Transformer-XL and Compressive Transformers propagate caches or compressed representations between segments \citep{dai2019transformerxl,rae2020compressive}, and Infini-attention combines local attention with a compressive associative state \citep{munkhdalai2024infini}. SSA shares the fixed inference-state objective, but keeps its long-range memory as selected key--value chunks read by softmax attention, so storage is exact as well as lossy; Transformer-XL also keeps exact representations, but only of the most recent segment, whereas SSA keeps a learned subset.

\paragraph{Cache selection and learned eviction.} H$_2$O, TOVA, Scissorhands, SnapKV, and StreamingLLM reduce the cache of a pretrained Transformer using attention statistics or positional rules \citep{zhang2023h2o,oren2024tova,liu2023scissorhands,li2024snapkv,xiao2024streamingllm}. We compare with four of them, and with Dynamic Memory Sparsification \citep{nawrot2025dms} and Dynamic Context Pruning \citep{anagnostidis2023dynamic}, at the 400M scale (\S\ref{sec:scaling_posttraining}). Dynamic Memory Compression learns to compress caches through continued training \citep{nawrot2024dmc}. Expire-Span \citep{sukhbaatar2021expire} and Selective Attention \citep{leviathan2025selective} learn which memories to drop, and Landmark Attention \citep{mohtashami2023landmark} learns to retrieve blocks from a full cache. Learned token eviction has also been studied in hybrid linear and sparse-attention models \citep{he2025lte}. HOLA combines a recurrent delta-rule state with a bounded exact cache whose retention rule uses the magnitude of the recurrent update \citep{cui2026hola}, and IndexMem combines a learned cache indexer with latent memory for evicted content \citep{yang2026indexmem}. In SSA the selected attention memory is the sole long-range memory, and its reader and retention policy are trained jointly from scratch. We do not claim that learned eviction is new; we claim that our approach and training combination is. Unlike these methods, SSA uses a hard, fixed-size memory at inference and in hard-phase training, rescores every resident against the new candidate at every eviction, and adds counterfactual credit for evictions through recovery. Sparse-reading methods such as Quest and Native Sparse Attention \citep{tang2024quest,yuan2025nsa} reduce how much history a query reads without bounding what is stored; SSA, like SSMs, bounds both (App.~\ref{app:related}).

\section{State Space Attention}
\label{sec:arch}

\subsection{A bounded attention state}

\begin{figure}[t]
    \centering
    \includegraphics[width=0.6\linewidth]{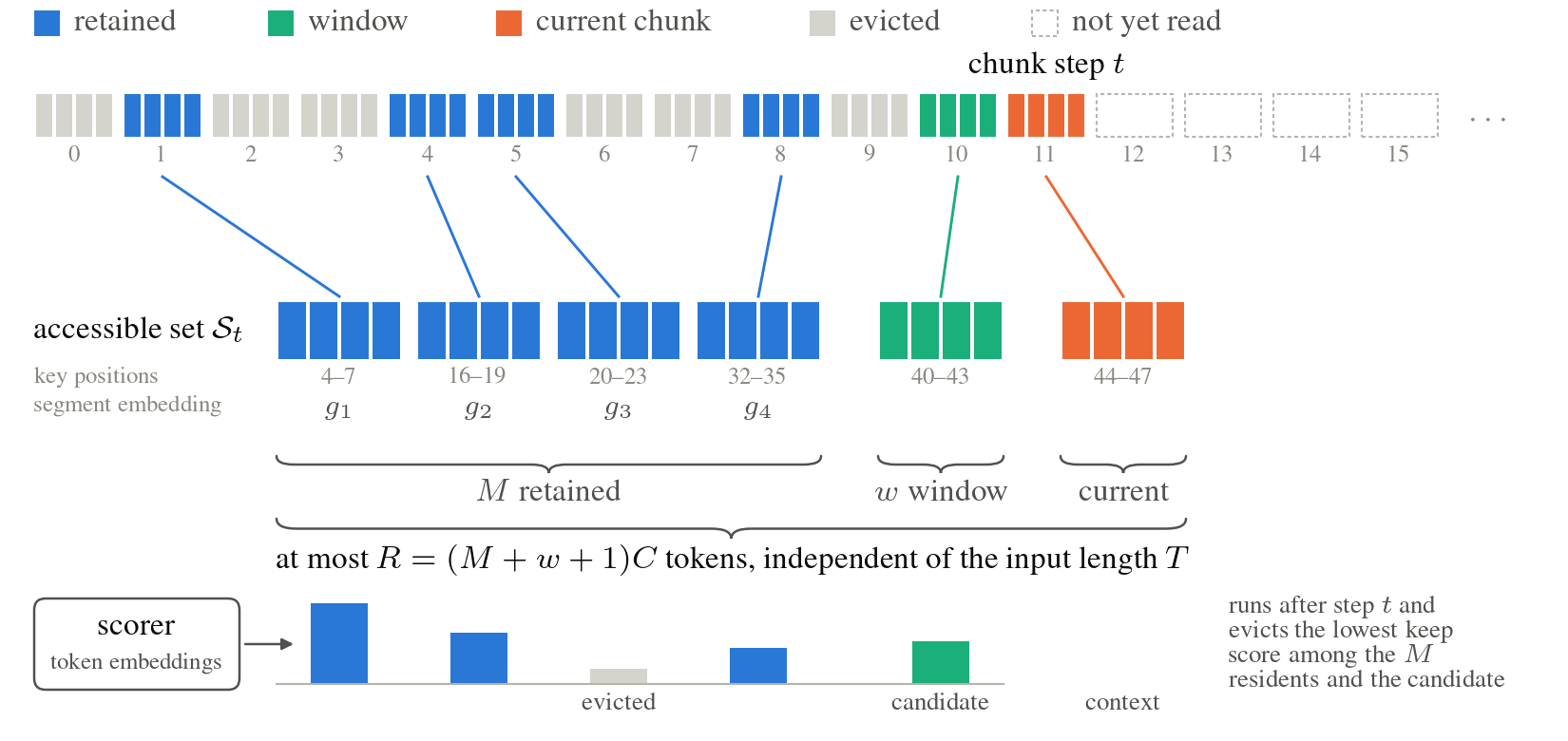}
    \caption{The bounded attention state at chunk step $t$, for a toy configuration with $M=4$, $w=1$, and $C=4$. Top: the input, with each chunk colored by its role. Middle: the accessible set $\mathcal S_t$, shared by all layers and heads. Bottom: the scorer rescores the candidate and the $M$ residents and evicts the lowest-scoring chunk (here chunk 5).}
    \label{fig:boundedstate}
\end{figure}

Let an input of length $T$ be divided into chunks of $C$ tokens. At chunk step $t$, SSA
maintains at most $M$ retained chunks and a local window of \(w\) preceding chunks (Figure~\ref{fig:boundedstate}). Including
the current chunk, the reader has access to at most
\begin{equation}
R=(M+w+1)C
\label{eq:capacity}
\end{equation}
tokens, independent of $T$. Chunk membership is shared across layers and heads in the basic recipe (but can also be layer or head specific), and each layer stores its own keys and values for the held chunks. Queries use ordinary causal
attention over the accessible set (the set that contains the $R$ tokens) $\mathcal S_t$:
\begin{equation}
 p_{qk}=\frac{\exp(a_{qk})}{\sum_{j\in\mathcal S_t}\exp(a_{qj})},
 \qquad o_q=\sum_{k\in\mathcal S_t}p_{qk}v_k,
 \qquad a_{qk}=\frac{q^\top k+q^\top g_{\mathrm{slot}(k)}}{\sqrt{d_h}}.
\label{eq:read}
\end{equation}
Here $d_h$ is the head dimension and $g$ is a learned segment embedding for memory slots.
Keys preserve their original rotary positions \citep{su2024roformer}, and the current
chunk is masked causally. Empty slots are excluded from the softmax.

When a chunk leaves the recent local window, it becomes a candidate for long-range memory.
If memory is full, the scorer rescores all resident chunks and the new candidate and removes
the lowest-scoring member. Thus, the candidate itself can be removed, leaving the residents
unchanged. Rescoring the residents at every later decision lets retention respond to newly observed context.

\begin{figure}[t]
    \centering
    \includegraphics[width=0.65\linewidth]{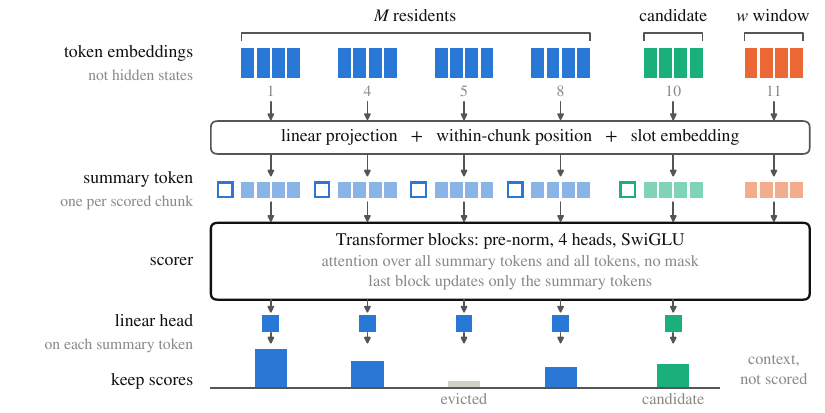}
    \caption{The scorer at one eviction decision, in the toy configuration of Figure~\ref{fig:boundedstate}. Each resident and the candidate has a summary token, which a linear head maps to a keep score; the lowest is evicted.}
    \label{fig:scorer}
\end{figure}

\paragraph{The scorer.}
The scorer is a small Transformer over the reader's token embeddings, not its hidden states (Figure~\ref{fig:scorer}); a scorer that read hidden states gave no gain for a large drop in training speed (App.~\ref{appendix:hiddenSSA}). It reads the tokens of the $M$ residents, the candidate and the $w$ window chunks, each with learned position and slot embeddings, and gives each resident and the candidate a learned summary token. Transformer blocks attend over all of them without a mask, a linear head maps each summary token to one keep score, and the lowest-scoring of the $M+1$ chunks is evicted. One set of scores is shared by all layers and heads of the reader. The scorer is small (App.~\ref{app:scorer}): 0.87M parameters beside a 3.18M reader in the 3M family, and 13.8M (3.5\% of the model) at 400M.

For $L$ layers of width $d$, the reader stores at most $2LdR$ key--value elements per
sequence. Token identifiers, positions, and scorer metadata also have bounded size.
Persistent inference memory is therefore constant in sequence length, and the attention work per token is bounded. Training activation memory is not constant, as the keys and values of evicted chunks must be held for backpropagation; in hard-phase training we measure memory use that is close to that of a standard Transformer (within 6\% from 2K to 64K context for our 400M configuration; App.~\ref{app:compute}). Evicted keys and values cannot be reread, though information from them can remain in the retained, contextualized keys and values of deeper layers (\S\ref{sec:lossy}); recovery (\S\ref{sec:training}) gives the scorer a training signal for them, and we propose that it can also be deployed during inference to recover lost information.

\subsection{Learning discrete retention}
\label{sec:training}

A hard eviction changes which information later predictions can use, and its effect can be delayed and discontinuous: a chunk that was unnecessary for steps 1 to 100 may become necessary at step 101, so the model must learn to keep it through those steps. We train the scorer with an optional continuous relaxation early in training and, in hard training, a straight-through gradient for hard reads along with counterfactual feedback for discarded chunks.

\paragraph{Soft retention warmup.}
Let $s_c^{(e)}$ be the keep score for chunk $c$ at eviction event $e$. Negated scores
define drop probabilities $p^{(e)}=\softmaxop(-s^{(e)}/\tau)$. During warmup, cumulative
survival weights reduce the contribution of chunks to the attention calculation, rather than removing them:
\begin{equation}
 \omega_c^{(e)}=\prod_{e'\in H_c,\,e'\le e}(1-\lambda p_c^{(e')}),
 \qquad a_{qk}^{\mathrm{soft}}=a_{qk}+\log\omega_{c(k)}^{(e)}.
\label{eq:soft}
\end{equation}
The softmax of \eqref{eq:read} then runs over the keys of all preceding chunks, including those the hard policy would have evicted, so each chunk's attention weight is multiplied by its survival weight before renormalizing. $H_c$ contains the events in which $c$ participates. Increasing the eviction strength $\lambda$ and lowering the temperature $\tau$ sharpens selection. The relaxation lets useful chunks receive credit before retention becomes discrete, and it holds extra representations only during training: every reported evaluation, including those logged during the soft phase, runs in hard mode.

A down-weighted chunk can still be read, so the backbone can also correct the scorer: a large attention weight on the chunk signals that it should have been kept. The soft phase is optional: in our 400M ablations, runs that skipped it learned retrieval at least as reliably, and since it also needs more memory at long contexts, the ablations point to removing the soft signal from the post-training pipeline (\S\ref{sec:ablations}). It has not been ablated in pretraining; recovery has (Table~\ref{tab:scaling}), and our post-training results use the no-recovery model, a worst case since pretraining with recovery gave the better NLL.
\looseness=-1

\paragraph{Hard retention and read gradients.}
Hard training executes discrete evictions, with stochastic selection from the drop scores for
exploration, while evaluation uses the lowest keep score as the chunk to evict. A centered straight-through
estimator \citep{bengio2013ste} preserves the hard forward read and differentiates a
normalized read weight in the backward pass. The local derivative for a chunk is
\begin{equation}
 \frac{\partial\mathcal L}{\partial\log\omega_c}
 =\sum_q\sum_{k\in c}p_{qk}\,g_q^\top(v_k-o_q),
 \qquad g_q=\frac{\partial\mathcal L}{\partial o_q}.
\label{eq:advantage}
\end{equation}
This measures a chunk's contribution relative to the current attention output: it gives credit through surviving reads, but cannot evaluate a chunk that is no longer available (derivation in App.~\ref{app:gradients}).

\paragraph{Counterfactual recovery during training.} For selected training failures, recovery replays the example while forcing a previously evicted chunk to stay available, with the rest of the policy acting under the same capacity constraint. If the prediction then passes the acceptance criterion, the scorer receives survival credit and a margin loss favoring that chunk over its replacement (App.~\ref{app:gradients}). This supplies feedback exactly where read gradients are absent. Recovery searches all discarded chunks; we also provide a cheaper search that ranks candidates with the method of Quest \citep{tang2024quest} (0.73 AUC against the exhaustive search), which our experiments do not use. Recovery is off at inference by default, but can serve as a deployment knob: its thoroughness interpolates between constant memory, with possible irretrievable forgetting, and quadratic memory, where the model retries using every evicted chunk; at inference the model's confidence is used in place of the loss.

\paragraph{Soft credit approximates counterfactual credit.}
Let $A_c=-\partial\mathcal L/\partial\log\omega_c$ (\eqref{eq:advantage}). With the eviction trajectory fixed, the soft model's scorer gradient is the gradient of a credit-weighted survival loss,
\begin{equation}
 \frac{\partial\mathcal L}{\partial\theta}\Big|_{\omega}
 =\frac{\partial}{\partial\theta}\sum_c\sg(A_c)\sum_{e\in H_c}\big[-\log(1-\lambda p^{(e)}_c)\big].
\label{eq:master}
\end{equation}
At $\lambda=1$, the survival term of the recovery loss (App.~\ref{app:gradients}) is this same loss with $A_c$ replaced by $\mathbf 1[c\text{ accepted}]\,w_c$, $w_c\geq 0$, and the pairwise margin is its two-point, victim-only version. Soft training and recovery therefore train the scorer through the same loss and differ only in the credit that chunk $c$ receives. Recovery measures that credit directly, as the loss change when $c$ is kept and another chunk is evicted instead, which is the quantity hard retention depends on. $A_c$ is its local, first-order counterpart: it holds membership and all other read weights fixed. As $\tau\to0$, $\omega_c\to\{0,1\}$ and the soft gradient fades for chunks that are already evicted, exactly where recovery still measures value. The soft phase is thus an approximation of recovery that works only while retention is soft; in our 400M ablations, runs without it learned retrieval in 2 of 2 runs, while neither run without recovery did (\S\ref{sec:ablations}).

\subsection{Training the reader and executing the policy}
\label{sec:recipe}

For retrieval over prose, we first train next-character prediction and then introduce retrieval tasks of increasing difficulty, so that the reader develops before it must learn query selection. The multi-key variable tracking (MKVT) experiments (\S\ref{sec:mkvt}) examine the language phase, and the 400M ablations (\S\ref{sec:ablations}) the soft phase, the straight-through estimator and recovery. By default the soft phase is on and recovery off in language pretraining, and both are on in task-specific post-training; the MKVT suite and one 400M model also test recovery during language training (Tables~\ref{tab:mkvt_ssa} and~\ref{tab:scaling}). We hypothesized that language modeling alone leaves the scorer data-starved. The 400M pretrained scorer is consistent with this: on 8K inputs it applies a fixed positional rule (\S\ref{scaling:pretraining}) and becomes selective only after long-context post-training (App.~\ref{app:lm_diagnostics}).

Training does not run the reader sequentially. The scorer's features do not depend on reader activations, so the only sequential part is a sweep of the small scorer, run without gradients, that records the retained sets; all of the reader's attention is then computed in parallel over the whole sequence (App.~\ref{app:gradients}). SSA still trains more slowly per update than a Transformer (\S\ref{sec:memory}).

\section{Experiments}
\label{sec:experiments}

We evaluate memory selection, reasoning, and optimization separately. Long-context cued copying tests whether relevant content can survive beyond a recent window. Multi-hop induction tests the reader's ability to compose information stored in its state. MKVT (Multi-Key Variable Tracking) combines prose with query-dependent retrieval and exposes the interaction between pretraining and retention. We compare SSA with full-attention Transformers, Gated DeltaNet (GDN) \citep{yang2025gateddeltanet}, HOLA \citep{cui2026hola}, laLTE \citep{he2025lte}, and fixed eviction policies where available. GDN is our SSM baseline because GDN and its variants are widely used in recent deployed models \citep{qwen2025qwen3next,kimi2025linear,qwen2026qwen35,qiu2026qwen38next}; GDN2 \citep{hatamizadeh2026gdn2} appeared during this work, so we did not include it. HOLA is a recent preprint that addresses the recall weakness of SSMs with a bounded exact cache, and laLTE learns token eviction in a hybrid linear-and-sparse-attention model.

Protocols are in Appendices~\ref{app:protocols} and~\ref{app:post-training}. Cued copying and 400M MKVT report greedy exact match; at 400M, MQAR is mean query accuracy, RULER is a mean of task scores, and $k$-hop is generated token accuracy excluding the null symbol; the 20M multi-hop and the small-scale MKVT comparisons are teacher-forced. We compare against two Transformers: a backbone-matched one, with SSA's reader dimensions and no scorer, and a larger one that matches SSA's total parameter count. Unless noted, SSA shares chunk membership across layers and heads, with a readable budget of $R=128$ tokens in the 3M suite ($M=14$, $C=8$, $w=1$, FP32 keys and values), $R=256$ in the 20M suite ($M=30$), and $R=576$ in the 400M runs ($M=34$, $C=16$); layer and head-wise membership (implemented in our code), BF16 storage, and GDN hybrids are compared in Tables~\ref{tab:ssa_variants} and~\ref{tab:ssa_hybrids}. We keep SSA's parameter count and state size in bytes as close to each baseline's as possible, and its reader is never larger than the Transformer's. SSA's learning errors and optimization problems closely follow the Transformer's (Apps.~\ref{app:protocols} and~\ref{app:mkvt}); learned retention adds a few failure points of its own, which some configurations of our training recipe address (\S\ref{sec:ablations}).

In cued copying, where a marked payload must be reproduced from a 3072-token sequence, 24 times SSA's 128-token readable state, SSA reaches 98.70\% against 100\% for full attention, while the fixed eviction rules fail with the same reader and capacity (0.00\% for drop-oldest, 2.86\% for random eviction; Table~\ref{tab:cued}, App.~\ref{app:protocols}). GDN, HOLA, and laLTE also reach 100\%, since the content between payload and query is noise that a recurrent state need not track; the task is meant to establish the feasibility of bounded retention, not an advantage over recurrent models. We stress test cued copying with information composition at 400M scale also. 

\subsection{Preserving multi-hop composition}
\label{sec:khop}

The $k$-hop induction task composes previous-occurrence lookups in a random string \citep{sanford2024transformers}; we sample the hop count from zero through sixteen and predict the target sequence as a suffix. We use the 20M family for the preliminary experiments, as the 3M family is too small for the task (Table~\ref{tab:khop_3m}). The whole sequence fits SSA's 256-token readable state, so this tests the bounded reader, not retention under eviction.

With fresh data SSA matches both Transformer controls (99.65\% against 99.75\% and 99.55\%). With a fixed training set it is 24.3 points below the same-backbone Transformer (70.05\% against 94.35\%). Their read mechanisms are nearly the same, so we believe this gap comes from optimization, which we did not test. GDN and HOLA are far less accurate (34.40\% and 73.40\% with fresh data, 0.10\% and 6.50\% with a fixed training set; Table~\ref{tab:khop} in App.~\ref{app:protocols}) although their inference states are about the size of SSA's (\S\ref{sec:memory}): how memory is represented and used, and not only its size, matters for tasks that need exact access to earlier context \citep{jelassi2024repeat}. This gap is further widened at our larger scale runs. 
\subsection{Learning query-dependent retrieval in prose}
\label{sec:mkvt}

MKVT is a character-level adaptation of RULER's multi-key needle-in-a-haystack task \citep{hsieh2024ruler}, with key--value statements embedded in TinyStories prose \citep{eldan2023tinystories} and distractor bindings that force the model to use the query. We evaluate four tasks at a 1024-character context, varying the number of bindings and eligible query keys, with three seeds per task (App.~\ref{app:mkvt}); a case counts as solved when fixed-target whole-answer exact match reaches 99\%.

\paragraph{Retrieval learning can fail without memory loss.}
Across 24 Transformer recipes, failed checkpoints often produced a candidate value without responding to the query, even though the Transformer retains the full context; this matches the delayed emergence of attention operations in multi-step tasks \citep{hoffmann2024eureka}. Extending the language phase solves most development cases (Table~\ref{tab:mkvt_pretrain}); details and controls are in App.~\ref{app:mkvt}.

\paragraph{SSA separates successful reading from successful retention.} With extended language pretraining, 44 of 48 SSA runs reach 90\% first-stage answer exact match within five retrieval epochs, and at the end of training 40/48 reach $\geq99\%$ fixed-target whole-answer exact match and 46/48 reach $\geq97.5\%$ (Table~\ref{tab:mkvt_ssa}). The reader answers accurately when a chunk containing part of the queried statement is readable at the query; the remaining errors are concentrated where all statement chunks have been evicted, which associates the residual gap with retention.

\label{sec:lossy}
Information can also reach the answer indirectly, through retained chunks computed while the source was still readable. We call an example $n$-hop when the shortest such path has $n$ steps: 0-hop means the source chunk is in memory at the query, 1-hop means a chunk in memory was processed while the source was in memory. This is the lossy path of \S\ref{sec:arch}. In our data, 2.9\% and 1.3\% of examples are 1-hop and 2-hop, and SSA answers 63.1\% and 46.8\% of them exactly, indicating this as a real but small effect (App.~\ref{app:mkvt}).

\subsection{Scaling experiments}
\label{sec:scaling}

\subsubsection{Pretraining}
\label{scaling:pretraining}
We pretrain eight approximately 400M-parameter models on 15B SlimPajama~\citep{cerebras2023slimpajama} tokens, a relatively cheap scale still in the range of small language models (SLMs) \citep{liu2024mobilellm,lu2024slmsurvey}: Transformers at 576, 2K, and 8K context; GDN; HOLA; and three SSA variants (the original scorer without recovery, the original scorer with recovery during pretraining, and a larger four-layer, width-896 scorer without recovery, whose parameters come from the reader's feed-forward layers). All readers have 24 layers and width 1024, and all SSA variants are pretrained at 2K with a readable budget of $R=576$ tokens.

At 2K, the SSA variants reach 2.650--2.665 nats per token (Table~\ref{tab:scaling}): slightly worse than GDN, HOLA, and the longer-context Transformers, but better than Transformer-$576$. At 8K, the original SSA reaches 4.334 nats, against 5.039 for the 2K-pretrained Transformer with unrestricted attention, while Transformer-$576$ stays at 2.686. Recovery during pretraining improves SSA's 8K NLL to 3.626 and a larger scorer to 4.191, with little change at 2K; neither removes the degradation. 

The degradation comes from the old chunks that the pretrained policy keeps. On 8K inputs the no-recovery model applies a fixed rule: it holds 15 early chunks for the whole sequence and fills its remaining 19 slots with the most recent chunks. Replacing this rule at evaluation with a sliding window over the 34 most recent chunks, with the same weights, gives 2.697 nats at 8K, level with Transformer-$576$ (2.686). Keeping just the first chunk and the 33 most recent gives 3.739 nats at 8K but no loss at 2K, which points to keys beyond the 2K pretraining distances (App.~\ref{app:lm_diagnostics}).

\begin{table}[h!]
\centering
\small
\caption{ Held-out SlimPajama negative log-likelihood after 15B-token pretraining, evaluated at 2K and 8K segment lengths. Results use one seed; lower is better.
 }
\label{tab:scaling}
\begin{tabular}{@{}lrrr@{}}
\toprule
Model & Parameters & NLL at 2K & NLL at 8K \\
\midrule
 Transformer (576) & 400.1M & 2.695 & 2.686 \\
 Transformer (2K)  & 400.1M & 2.615 & 5.039 \\
 Transformer (8K)  & 400.1M & 2.623 & 2.557 \\
 Gated DeltaNet    & 399.9M & 2.590 & 2.549 \\
 HOLA             & 399.8M & 2.591 & 2.558 \\
 \midrule
 SSA, no recovery   & 399.8M & 2.656 & 4.334 \\
 SSA, recovery      & 399.8M & 2.650 & 3.626 \\
 SSA, larger scorer & 400.0M & 2.665 & 4.191 \\
\bottomrule
\end{tabular}
\end{table}

GDN's lead at 2K is in line with prior comparisons at this scale on SlimPajama \citep{yang2024deltanet}.

\subsubsection{Post-training at 400M}
\label{sec:scaling_posttraining}

We post-train the 400M models on MKVT, atomic MQAR, $k$-hop reasoning, and RULER, with equal weighting across the four task families, and equal weighting across the 13 RULER subtasks (App.~\ref{app:post-training}). Each model in Table~\ref{tab:posttraining_final} receives 7,800 mixed-task updates and then 1,760 updates on higher-hop reasoning with $N=256$ and $h\in\{4,8,16\}$, with primary training examples and reader learning-rate schedules matched across models (the matched schedule gave the best overall results among those we tried). For SSA we also compare training the scorer throughout $k$-hop specialization with freezing it from the mixed-task checkpoint, as the $k$-hop state fits within SSA's context window.
	
\paragraph{Learned retention at a fixed attention budget.} With a readable budget of $R=576$ tokens, SSA with a trainable scorer reaches 96.57\% on MKVT and a 56.40\% mean RULER score at 8K, against 8.13\% and 14.61\% for Transformer-$576$. A 2K-pretrained Transformer restricted to a 576-token window, which separates the window from the pretraining context, reaches 7.93\% and 15.76\%, and with a 2K window it reaches 34.73\% and 36.50\%. Learned retention improves retrieval over recency windows of the same budget and of about 4 times the budget. Full 8K attention remains stronger: the 2K and 8K-pretrained Transformers with an 8K window reach 99.60\% and 99.90\% on MKVT, within 3.33 points of SSA, and 70.72\% and 76.86\% on RULER, where the gap is larger.

\paragraph{Multi-hop composition.} The $k$-hop sequence fits within SSA's 576-token readable budget, so this tests multi-hop computation without eviction pressure. We report generated token accuracy excluding the null symbol (we do not discuss whole-sequence exact match, which is highly unstable and depends on run-to-run randomness for 256-token answers; App.~\ref{app:post-training} reports it). After the 1,760 dedicated updates, SSA with a trainable scorer reaches 99.984\%, 99.971\%, and 99.536\% at 4, 8, and 16 hops, against 99.997\%, 99.976\%, and 99.457\% for the 8K-pretrained Transformer and 99.997\%, 99.970\%, and 99.472\% for Transformer-$576$; freezing the SSA scorer gives 99.609\% at 16 hops. These four models are within 0.2 points of each other at every hop count. GDN and HOLA stay at chance (about 26\%). The 2K-pretrained Transformers are far less accurate on $k$-hop than the others under the shared schedule (48.251\%, 25.936\% and 26.068\% at 16 hops with 576, 2K and 8K windows); the two with 2K and 8K windows saturate the task at a higher learning rate, at a cost in perplexity and on the other tasks, so we view this as an optimization gap (App.~\ref{app:posttraining_forgetting}). Overall SSA combines this reasoning performance with strong MKVT and MQAR retention under a 576-token budget, while remaining behind the 8K-pretrained Transformer on RULER and perplexity; freezing the scorer changes MKVT by a few points and leaves RULER and perplexity nearly unchanged (Table~\ref{tab:posttraining_final}, App.~\ref{app:posttraining_forgetting}).

\begin{table*}[t]
\centering
\caption{
Final 400M post-training results. Each $k$-hop entry reports generated token accuracy excluding the null symbol (\%). Transformer labels denote pretraining context $\rightarrow$ task-attention window. Metric definitions, training details, and forgetting results are provided in Appendix~\ref{app:post-training}.}
\label{tab:posttraining_final}
\setlength{\tabcolsep}{3pt}
\renewcommand{\arraystretch}{1.12}
\resizebox{\textwidth}{!}{%
\begin{tabular}{lrrrrrrrrrrr}
\toprule
& \multicolumn{3}{c}{MKVT}
& \multicolumn{1}{c}{MQAR}
& \multicolumn{2}{c}{RULER}
& \multicolumn{3}{c}{$k$-hop token accuracy}
& \multicolumn{2}{c}{PPL $\downarrow$} \\
\cmidrule(lr){2-4}
\cmidrule(lr){5-5}
\cmidrule(lr){6-7}
\cmidrule(lr){8-10}
\cmidrule(lr){11-12}
Model
& 2K & 4K & 8K
& Query
& 4K & 8K
& $h=4$ & $h=8$ & $h=16$
& 2K & 8K \\
\midrule
SSA, frozen scorer
& 98.10 & 98.13 & 97.53 & 99.954
& 64.14 & 56.29
& 99.984 & 99.968 & 99.609
& 22.99 & 27.00 \\
SSA, trainable scorer
& 96.13 & 95.97 & 96.57 & 99.952
& 64.00 & 56.40
& 99.984 & 99.971 & 99.536
& 23.07 & 26.91 \\
\midrule
Transformer $576\rightarrow576$
& 28.90 & 14.67 & 8.13 & 85.739
& 22.04 & 14.61
& 99.997 & 99.970 & 99.472
& 20.34 & 19.79 \\
Transformer $2\mathrm{K}\rightarrow576$
& 29.20 & 14.53 & 7.93 & 85.829
& 22.70 & 15.76
& 99.982 & 99.721 & 48.251
& 21.49 & 20.77 \\
Transformer $2\mathrm{K}\rightarrow2\mathrm{K}$
& 99.90 & 66.57 & 34.73 & 98.162
& 60.56 & 36.50
& 99.764 & 34.184 & 25.936
& 19.72 & 19.03 \\
Transformer $2\mathrm{K}\rightarrow8\mathrm{K}$
& 99.97 & 99.97 & 99.60 & 99.992
& 78.27 & 70.72
& 82.526 & 29.424 & 26.068
& 19.73 & 22.28 \\
Transformer $8\mathrm{K}\rightarrow8\mathrm{K}$
& 100.00 & 100.00 & 99.90 & 99.976
& 77.58 & 76.86
& 99.997 & 99.976 & 99.457
& 20.51 & 18.59 \\
\midrule
GDN
& 99.80 & 99.60 & 99.50 & 99.965
& 69.07 & 68.91
& 26.497 & 26.184 & 25.594
& 18.77 & 18.57 \\
HOLA
& 99.30 & 99.53 & 99.53 & 99.882
& 62.39 & 58.04
& 27.332 & 26.167 & 25.847
& 19.14 & 18.84 \\
\bottomrule
\end{tabular}%
}
\end{table*}

\paragraph{Composition under sustained eviction.}
To test composition when the source must survive eviction, we hide the $k$-hop source block in an 8K input and ask the question at the end (Table~\ref{tab:haystack}, App.~\ref{app:haystack}). We again compare token accuracy excluding the null symbol, where chance is about 25\%. With the block at the start and training from each model's $k$-hop checkpoint, SSA reaches 99.2\% against 99.9\% for the full-attention 8K Transformer, while Transformer-$576$, GDN and HOLA stay at chance (22.6\% to 27.2\%). From the mixed-task checkpoint, SSA reaches 77.4\% after 1,760 updates (100.0\%, 99.9\% and 25.9\% at 4, 8 and 16 hops) and the other models stay at chance (22.5\% to 26.7\%); after 3,520 updates SSA reaches 85.2\% and the full-attention Transformer 53.1\%. With the block at a random position every model is at chance, the full-attention Transformer included; SSA's memory still holds 97.1\% of the block's chunks there, so the failure is in using the block, not in retaining it.

\paragraph{Training-free cache eviction.} At SSA's 576-token budget, cache-eviction methods that select without seeing the question fail on MKVT at 8K on the 2K-pretrained Transformer: StreamingLLM reaches 7.6\%, H$_2$O 5.3\% and TOVA 6.2\%, against 96.6\% for SSA (RULER: 3.8 to 11.6 against 56.4). SnapKV, which reads the whole prompt including the question before selecting, reaches 99.1\% on MKVT and 51.0 on RULER (Table~\ref{tab:eviction}, App.~\ref{app:eviction}). These methods were not trained under eviction, so the comparison is learned against rule-based retention at an equal budget.

\paragraph{Learned eviction.} We also train Dynamic Memory Sparsification \citep[DMS;][]{nawrot2025dms} from the 2K-pretrained Transformer through the same 9,560 post-training updates, with the hyperparameters of its paper and official code (no search of our own) and a compression ratio that matches SSA's 576 tokens at 8K (Table~\ref{tab:dms}, App.~\ref{app:eviction}). DMS chooses separately for each layer and head: each head keeps 344 tokens on average against SSA's 576, but across its 96 layer and head pairs DMS holds keys and values from about 4,480 distinct tokens, where SSA holds the same 576 at every layer and head; we acknowledge this as an advantage of DMS, and SSA can be extended the same way. On MKVT at 2K and 4K the two are nearly equivalent, with DMS a few percentage points ahead; they are also nearly equivalent on MQAR, and DMS has lower perplexity (20.92 against 23.07 at 2K). On MKVT at 8K they are nearly equivalent, with SSA leading by one percentage point (96.57\% against 95.43\%). SSA is better on RULER at 8K (56.40 against 50.84) and especially on $k$-hop, where token accuracy at 4, 8 and 16 hops is 99.98\%, 99.97\% and 99.54\% against 99.95\%, 97.55\% and 38.61\%. We also train Dynamic Context Pruning \citep[DCP;][]{anagnostidis2023dynamic} the same way, with its sparsity weight set by a rule fixed before training. It keeps far fewer tokens than the budget allows, 62 per layer at the end of 8K MKVT prompts, and it is far worse than DMS and SSA on every task: 23.9\%, 17.9\% and 8.6\% on MKVT at 2K, 4K and 8K, 50.4\% on MQAR, no exact answers and about 25\% token accuracy on $k$-hop, and a perplexity of 56.8 at 2K (Table~\ref{tab:dms}, App.~\ref{app:eviction}).

\paragraph{Component ablations.} \label{sec:ablations} Starting from the same pretrained SSA, we post-train for 1,560 updates with one component removed at a time, two runs per arm (Table~\ref{tab:ablations}, App.~\ref{app:ablations}). The outcome is bimodal: a run either learns to match the query key (64\% to 98\% on MKVT 8K) or stays near 8\%. Without recovery neither run learns; the full recipe learns in 1 of 3 runs; hard-only training, which replaces the soft phase with further hard updates and recovery passes, learns in 2 of 2, as do both scorers that score each chunk once; without the straight-through estimator one of two runs learns in part. With so few runs we cannot rank the arms, and we do not claim that the soft phase or chunk rescoring speeds up retrieval learning; recovery seems to be essential, since no run without recovery learned. The same compute appears better spent on recovery.

\paragraph{Scorer capacity and recovery allocation.} A larger scorer (four layers of width 896, 13\% of the parameters) reaches 99.87\% on 8K MKVT after 7,800 mixed-task updates, against 98.40\% for the original, with the same RULER mean and worse perplexity (22.05 against 21.40 at 2K), so we keep the small scorer. Pilots that reserve recovery slots per task family differ between reruns with the same seed, and we draw no conclusion from them (App.~\ref{app:posttraining_scorer_capacity}).

\paragraph{Language modeling and the local window.} After post-training, SSA has the highest perplexity in Table~\ref{tab:posttraining_final} (23.07 at 2K and 26.91 at 8K with a trainable scorer). Most of this excess comes from its one-chunk local window rather than from keeping a memory. The post-trained scorer keeps old chunks for retrieval: its residents are spread over the whole input and it evicts the incoming chunk in 66\% of decisions (App.~\ref{app:lm_diagnostics}), so a query reads little recent text. With the same weights, replacing the 12 lowest-scored of its 34 memory chunks with the 12 most recent chunks, which widens the window to $w=13$, lowers perplexity from 23.07 to 20.95 at 2K and, with query--key distances capped at the pretraining length, from 24.92 to 20.59 at 8K, against 20.84 and 20.17 for a sliding window over all 35 chunks; widening the window further changes perplexity by at most 0.44 (Table~\ref{tab:window_size}). Content-blind memory shows the same knee near 13 chunks (208 tokens) at every budget we tried (App.~\ref{app:window_size}). This is consistent with the literature, which finds that language models draw most of their predictive signal from the last few hundred tokens and that perplexity changes little once the context passes this range \citep{khandelwal2018sharp,press2021shortformer}. Our 400M models use $w=1$, as this was the original config we trained the models on, but this is easily modifiable.

\subsection{Memory and computational cost}
\label{sec:memory}
For a standard multi-head Transformer with the same reader dimensions, cached keys and values take $2BLdTb$ bytes at sequence length $T$ ($B$ the batch size, $b$ the bytes per element); SSA replaces $T$ with the fixed capacity $R$. For the 20M family in FP32, SSA's state is a constant 6.00 MiB, slightly below the 6.11 MiB of GDN's and HOLA's recurrent matrices and convolution history, whereas a Transformer with SSA's reader dimensions needs 24 MiB at 1024 tokens and 240 MiB at 10,240. We measure memory in bytes as a neutral, downstream metric. SSA thus combines a state budget comparable to the recurrent baselines with an attention read over retained representations. Training costs more: the soft phase retains extra representations, parallel reads materialize activations across the example, and recovery repeats computations.

We also measure how SSA's cost scales with context length and model size against a full-attention Transformer of the same shape (Table~\ref{tab:cost_grid}), with SSA's readable budget fixed at 576 tokens and its scorer at 3.45\% of the parameters. At inference, as expected, SSA's decoding time per token does not grow with context: at 64K it is 0.24 to 0.56 of the Transformer's, with a constant key--value cache (54\,MiB against 6.0\,GiB at the 367M shape). SSA reads a prompt more slowly, since its eviction decisions are sequential, so its total time is lower only after a few hundred generated tokens. In training, SSA is slower in every cell: the hard-phase ratio falls with model size and, from 25M up, with context beyond 8K, and peak hard-phase memory is at most 1.57 times the Transformer's. Our 400M models have the 367M shape but 4 attention heads of dimension 256 in place of 16 of dimension 64, as do the GDN and HOLA models. On that layout the Transformer trains 1.7 to 4.0 times more slowly, so SSA's hard-phase ratio is lower there, from 2.8 at 2K to 0.68 at 64K (App.~\ref{app:compute}).

In our 400M training runs, an SSA update takes 3.6 (soft) and 6.0 (hard) times a Transformer's 2.5\,s in 2K pretraining, and 13 times with recovery. In post-training on inputs up to 8K the ratios are 3.0 in the soft phase, 2.3 in the hard phase, and 3.3 in the hard phase with recovery, which adds 45\% to a hard update (App.~\ref{app:compute}). The ratios therefore depend on context length and model size.

\section{Limitations}

All 400M results use a single seed for pretraining and post-training.
Our smaller-scale experiments show seed sensitivity on some tasks, and at 400M two reruns of a 1,560-update post-training pilot, with the same configuration and seed, ended near 30\% MKVT accuracy where the original run reached 90\%. Gaps of a few points between 400M rows, including the frozen versus trainable-scorer comparison, may therefore be noise. We also do not have a scaling-law analysis: the 400M models are one larger setting. We expect joint training of the scorer and the backbone may need more data than training a backbone with a fixed eviction policy, a limitation when data is scarce.

Our implementation trains SSA more slowly than a Transformer: 3.6 to 6.0 times per update in 400M pretraining without recovery, 13 times with it, and 2.3 to 3.3 times in post-training (\S\ref{sec:memory}, App.~\ref{app:compute}). At inference, at our 400M scale and above, SSA is faster per generated token than a Transformer, and its per-token time does not grow with context (Table~\ref{tab:cost_grid}), but it reads the prompt 2 to 6 times more slowly at this size, so it is ahead in total time only after a few hundred generated tokens. We do not show a better accuracy-per-cost trade-off than the baselines at 400M.

SSA's language modeling degrades beyond its 2K pretraining context, and after post-training SSA has the highest perplexity of all models in Table~\ref{tab:posttraining_final}, at 2K (22.99--23.07, against 18.77--21.49) and at 8K (26.91--27.00, against 18.57--22.28). The gap comes mostly from SSA's one-chunk local window: the learned policy fills memory with old chunks, so the reader sees little recent text. With the same weights, a 13-chunk window brings perplexity to 20.95 at 2K and 20.59 at 8K with distances capped (\S\ref{sec:scaling_posttraining}, App.~\ref{app:window_size}), and capping the distances at which old chunks are read removes about half of the rise from 2K to 8K (App.~\ref{app:lm_diagnostics}). Within our experiments, this does not translate to visibly worse text.

SSA trails the full-attention Transformers and GDN on RULER: at 8K it scores 56.29--56.40\%, against 76.86\% for the 8K-pretrained Transformer and 68.91\% for GDN; HOLA is within two points (58.04\%), and Transformer-576 is far below (14.61\%). It composes over a $k$-hop block hidden in an 8K input only with the block at the start; at a random position it stays at chance, as does full attention (\S\ref{sec:scaling_posttraining}). Tasks whose relevant information exceeds the memory budget remain untested, a problem for every bounded-memory architecture.

\section{Conclusion}

We introduce State Space Attention, which preserves strong retrieval and reasoning performance on the
studied tasks using a learned, bounded attention memory. Accurate contextual retrieval with a bounded cache needs three things: relevant information must fit within the state, the policy must learn to retain it, and the reader must learn to use the query; a failure on a task cannot by itself tell which one is missing. The training procedure of SSA addresses the last two. The results support learned selective memory as a route to contextual reasoning with constant inference state, and frame the eviction policy as a degree of freedom for Transformers: the standard fixed-context Transformer is the fixed first-in-first-out case, and a learned policy can take its place.

\section*{Acknowledgments}

C.P. is supported by an NSF CAREER Award (IIS-2239780), DARPA grants DIAL-FP-038 and AIQ-HR00112520041, and the Simons Collaboration on the Physics of Learning and Neural Computation. This work has been made possible in part by a gift from the Chan Zuckerberg Initiative Foundation to establish the Kempner Institute for the Study of Natural and Artificial Intelligence.

We thank William Tong for thoughtful conversations during the ideation stage.

\clearpage
\bibliography{iclr2027_conference}
\bibliographystyle{iclr2027_conference}
\clearpage
\appendix
\section{Training gradients and memory accounting}
\label{app:gradients}

\paragraph{The read-weight derivative.}
For one attention head, introduce a chunk bias $z_c=\log\omega_c$ into the logits of all
keys in chunk $c$. Let $P_{q,c}=\sum_{j\in c}p_{qj}$. Differentiating the softmax gives
\begin{equation}
 \frac{\partial p_{qk}}{\partial z_c}
 =p_{qk}\big(\mathbf{1}[k\in c]-P_{q,c}\big),
 \qquad
 \frac{\partial o_q}{\partial z_c}
 =\sum_{k\in c}p_{qk}(v_k-o_q).
\end{equation}
Contracting with the upstream gradient gives~\eqref{eq:advantage}; contributions
add across heads and layers.

During hard training, a straight-through weight has forward value one and a backward
path through the scorer's drop probability, for example,
$\widetilde\omega=1-p+\sg(p)$, which is the standard method of applying STE. Applying it inside a normalized attention read produces
the centered gradient above. Applying it after the softmax without renormalization
instead yields an uncentered contribution proportional to $g_q^\top v_k$. These are
different estimators, and neither can supply a direct read gradient for chunks absent from
the hard forward pass.

\paragraph{Counterfactual credit.}
Recovery runs only in the hard phase. At each step, the $\lfloor B/8\rfloor$ examples with the highest answer loss are buffered (eight for $B=64$). Every eight steps and at the end of each epoch, the buffer is replayed with the current scorer and reader, dropout off. For a buffered example, let $p^\ast$ be the answer position with the highest next-token loss and $W$ the answer targets among the $C$ positions starting at that failed prediction. For a membership record $\mathcal M$,
\begin{equation}
 \ell(\mathcal M)=\frac{1}{|W|}\sum_{i\in W}-\log p_\theta(x_i\mid x_{<i};\mathcal M).
\end{equation}
Candidates $\mathcal K$ are the chunks with index below $\lfloor p^\ast/C\rfloor-w$ that are not resident at chunk step $\lfloor p^\ast/C\rfloor$, i.e.\ everything discarded before the failure. For $c\in\mathcal K$, $\mathcal M_c$ is the recorded membership with one substitution: at the event that evicted $c$, and at any later event whose recorded victim is the chunk substituted out, the most droppable present chunk under the recorded scores is evicted instead. Let $\ell_0=\ell(\mathcal M)$, $\ell_c=\ell(\mathcal M_c)$ and $P=\{c\in\mathcal K:\ell_c<0.9\,\ell_0\}$. If $|\mathcal K|\ge8$ and $|P|>\max(6,\lfloor0.05|\mathcal K|\rfloor)$, the example is rejected, since a retention failure should have a specific cause. Otherwise, if $P\neq\emptyset$, $c^\ast=\arg\min_{c\in\mathcal K}\ell_c$ is accepted.
For an accepted $c^\ast$, let $H_{c^\ast}$ contain its recorded slate appearances, $e^\ast$ its eviction event, and $r^\ast$ the chunk evicted in its place. With $p^{(e)}=\softmaxop(-s^{(e)}/\tau_{\mathrm{credit}})$ and $\tau_{\mathrm{credit}}=1$,
\begin{equation}
 \mathcal L_{\mathrm{rec}}(c^\ast)=-\sum_{e\in H_{c^\ast}}\log(1-p_{c^\ast}^{(e)})+\softplusop\left(\frac{s_{r^\ast}^{(e^\ast)}-s_{c^\ast}^{(e^\ast)}}{\tau_{\mathrm{credit}}}\right).
\end{equation}
Each accepted example $b$ has relative improvement $\rho_b=\mathrm{clip}\big((\ell_0-\ell_{c^\ast_b})/\ell_0,0,1\big)$ and weight $w_b=\rho_b/\bar\rho$, with $\bar\rho$ the mean over examples accepted in the same replay. The replay adds $\frac{64}{N}\sum_b w_b\,\mathcal L_{\mathrm{rec}}(c^\ast_b)$ to the loss, where $N$ is the number of training examples since the previous replay. QUEST-style candidate ranking reduces search cost but does not replace this check~\citep{tang2024quest}.
For the 400M mixed-task refinement, global example selection is replaced by the equal-family recovery reservations described in Appendix~\ref{app:post-training}.

\paragraph{Parallel training.}
Training need not run the reader sequentially for every eviction. Because the scorer's features do not depend on reader activations, a sequential sweep by the scorer first records the retained sets; the recorded slates are then scored with gradients, and the reader processes the whole sequence in parallel subject to those sets. The sweep remains sequential, recovery adds task-dependent overhead, and the irregular read pattern rules out some efficient attention kernels, so SSA trains more slowly per update than a Transformer (\S\ref{sec:memory}); we believe optimization headroom remains.

\paragraph{Persistent state.}
For ordinary multi-head attention, SSA's reader stores $2BLdR$ floating-point elements.
The policy also stores bounded token identifiers and positions for the same retained
chunks. In the approximately 20M configuration, the reader uses six layers of width
512, eight-token chunks, thirty long-range chunks, and a one-chunk window, giving
$R=256$. FP32 reader state for batch one is 6 MiB; explicit token identifiers and compact
metadata add approximately 2.3 KiB. 

\section{Benchmark protocols and additional comparisons}
\label{app:protocols}

\paragraph{Cued copying.}
The task embeds two four-character payloads in a 3072-token context. Experiments use
2000 training examples, a maximum of 200 epochs, batch size 64, and a curriculum that
increases eviction pressure. The bounded models use a four-layer, width-256 backbone;
SSA uses fourteen retained chunks, one window chunk, and eight-token chunks, for a
128-token readable state. The Transformer comparison shares the reader backbone.
Table~\ref{tab:cued} reports the mean and seed range of completed runs at the learning rate with the better seed mean from
$\{10^{-4},3\cdot10^{-4}\}$; SSA uses shared membership with FP32 storage. laLTE runs use a small patch that allows training on the curriculum's short early stages (at most 64 tokens), which the released implementation for the paper rejects.

\begin{table}[h!]
\centering\small
\caption{Cued copying with a 3072-token context: greedy generation exact match (\%). Details in App.~\ref{app:protocols}.}
\label{tab:cued}
\begin{tabular}{@{}lrr@{}}
\toprule
Model / retention policy & Mean exact & Seed range \\
\midrule
Transformer (same backbone) & 100.00 & 100.00--100.00 \\
Gated DeltaNet & 100.00 & 100.00--100.00 \\
HOLA & 100.00 & 100.00--100.00 \\
laLTE & 100.00 & 100.00--100.00 \\
SSA & 98.70 & 97.66--100.00 \\
Drop oldest (SWA at matched memory) & 0.00 & 0.00--0.00 \\
Random eviction & 2.86 & 2.34--3.12 \\
\bottomrule
\end{tabular}
\end{table}

\paragraph{Multi-hop induction.}
The source string has length 100 over an alphabet of four symbols, with adjacent symbols
different. A hop finds the preceding occurrence of the current symbol and advances one
position; a missing predecessor produces a null target. Hop counts are sampled uniformly
from zero through sixteen. We follow the target recursion of
\citet{sanford2024transformers}, but place the complete target in an autoregressive
answer suffix instead of predicting targets at their original input positions. The
resulting sequence fits the 256-token state used in this experiment.

Training presents 8000 examples per epoch for 240 epochs, with batch size 64, gradient
clipping at one, and constant learning rate $10^{-4}$ or $3\cdot10^{-4}$. The fixed-data
condition reuses its training examples; the fresh-data condition resamples them each
epoch while keeping validation fixed. SSA and the backbone Transformer have six layers,
width 512, and feed-forward width 1376. The larger Transformer uses width 576 and
feed-forward width 1396. GDN and HOLA retain width 512 and feed-forward width 1921 from their source recipes.

The fixed-data condition uses no dropout.
In the fresh-data condition, SSA and the Transformers use
dropout 0.1, while GDN and HOLA retain dropout zero, since dropout did not appear to help them in our tests.
The remaining within-model training settings are unchanged.

For the 20M comparison, we report teacher-forced whole-sequence exact
match on each seed's 1,000-example validation set.
An example is correct only if the prediction at every one of the 100 answer positions
matches the reference, including null targets.
Learning-rate selection uses the mean of this metric over seeds. At small scales, HOLA appears substantially more accurate than GDN, but remains weaker than SSA.

At the 3M scale, no model learns the task under the same fixed-data protocol (Table~\ref{tab:khop_3m}). However, the larger Transformer ranks first, then SSA, then laLTE, then the smaller Transformer. Drop oldest is only slightly below SSA under teacher forcing, and slightly above it under greedy decoding.

\begin{table}[h!]
\centering\small
\caption{Multi-hop induction at the 3M scale ($N=100$, hop counts 0 to 16, fixed training set, no dropout): whole-sequence exact match (\%), teacher-forced and greedy, mean [seed range] over three seeds at the learning rate from $\{10^{-4},3\cdot10^{-4}\}$ with the better teacher-forced mean. The Transformer rows have a single run per learning rate. GDN + full attention is three GDN layers followed by one full-attention layer (Table~\ref{tab:ssa_hybrids}).}
\label{tab:khop_3m}
\begin{tabular}{@{}lrr@{}}
\toprule
Model & Teacher-forced & Greedy \\
\midrule
Transformer (same backbone) & 20.90 & 17.19 \\
Transformer (larger) & 30.30 & 28.91 \\
Gated DeltaNet & 0.00 [0.00--0.00] & 0.00 [0.00--0.00] \\
HOLA & 0.73 [0.00--2.20] & 0.52 [0.00--1.56] \\
laLTE & 25.47 [22.40--30.60] & 19.01 [17.19--22.66] \\
GDN + full attention & 16.60 [12.80--23.90] & 11.46 [6.25--18.75] \\
SSA (FP32, 14 chunks) & 27.13 [23.10--32.00] & 19.53 [16.41--24.22] \\
SSA (BF16, 30 chunks) & 26.57 [22.20--33.50] & 19.53 [16.41--25.00] \\
Drop oldest (FP32, 14 chunks) & 26.10 [24.10--29.40] & 21.09 [18.75--25.00] \\
Random eviction (FP32, 14 chunks) & 10.40 [7.40--13.20] & 5.73 [3.91--7.03] \\
\bottomrule
\end{tabular}
\end{table}

\begin{table}[h!]
\centering\small
\caption{Fresh-data multi-hop induction, both tested learning rates. Teacher-forced
whole-sequence exact match (\%), mean [range] of two completed seeds. These cells
underlie learning-rate selection in Table~\ref{tab:khop}.}
\label{tab:khop_lr}
\begin{tabular}{@{}lrr@{}}
\toprule
Model & $10^{-4}$ & $3\cdot10^{-4}$ \\
\midrule
Transformer (same backbone) & 99.75 [99.70--99.80] & 97.65 [96.00--99.30] \\
Transformer (larger) & 99.55 [99.40--99.70] & 99.25 [98.80--99.70] \\
Gated DeltaNet & 24.05 [20.90--27.20] & 34.40 [24.00--44.80] \\
HOLA & 70.30 [65.20--75.40] & 73.40 [70.50--76.30] \\
SSA & 99.65 [99.40--99.90] & 99.15 [98.90--99.40] \\
\bottomrule
\end{tabular}
\end{table}

\begin{table}[h!]
\centering\small
\caption{Multi-hop induction at 20M: teacher-forced whole-sequence exact match (\%). Fresh data resamples the training set every epoch; with fresh data, SSA and the Transformers also use dropout 0.1, and GDN and HOLA use none. Details in App.~\ref{app:protocols}. laLTE was not run at 20M; at 3M it slightly trails SSA on this task (Table~\ref{tab:khop_3m}).}
\label{tab:khop}
\begin{tabular}{@{}lrr@{}}
\toprule
Model & Fixed data & Fresh data, mean [seed range] \\
\midrule
Transformer (same backbone) & 94.35 & 99.75 [99.70--99.80] \\
Transformer (larger) & 82.75 & 99.55 [99.40--99.70] \\
Gated DeltaNet & 0.10 & 34.40 [24.00--44.80] \\
HOLA & 6.50 & 73.40 [70.50--76.30] \\
SSA & 70.05 & 99.65 [99.40--99.90] \\
\bottomrule
\end{tabular}
\end{table}

\paragraph{Additional retention tasks.}
Associative recall requires retrieving the value of a queried key among 64 bindings;
four bindings are eligible for querying. Copy join requires reproducing two payloads
in their specified order among noise, for more details please check our repository. Table~\ref{tab:retention} compares learned and fixed policies
with the same bounded reader. These experiments expose task dependence: learning
retention greatly improves cued copy, improves copy join  over dropping the oldest chunk, and associative recall remains
below the recurrent baselines' perfect accuracy in this suite.

\begin{table}[h!]
\centering\small
\caption{Additional retention comparisons: greedy generation exact match (\%),
mean of completed seeds (three; for copy join, learned retention has two at $10^{-4}$) at the better of two learning rates. Each task uses
identical reader capacity across the three policies.}
\label{tab:retention}
\begin{tabular}{@{}lrrr@{}}
\toprule
Task & Learned retention & Drop oldest & Random eviction \\
\midrule
Associative recall & 92.19 & 69.27 & 80.99 \\
Cued copy & 98.70 & 0.00 & 2.86 \\
Copy join & 100.00 & 36.46 & 94.53 \\
\bottomrule
\end{tabular}
\end{table}

Associative recall uses 32,000 training examples and 240 epochs; copy join uses 10,000
examples and 240 epochs for these three policies.

\paragraph{The 3M suite.} Each task uses three seeds per learning rate for cued copy, associative recall, alias
chains, $k$-hop and RULER variable tracking, two at $10^{-4}$ and three at $3\times10^{-4}$ for copy join, two for MKVT, variable tracking and
WikiText~\citep{merity2017pointer}, one or two for multi-query associative recall (MQAR) \citep{arora2023zoology}, and one for WikiText variable tracking and the $n$-gram suffix task. Alias chains,
MKVT, variable tracking and WikiText variable tracking use a single learning rate. Exact settings are all available in our repository.

\begin{table}[h!]
\centering\small
\caption{laLTE versus SSA (shared, FP32, 14 chunks) in the 3M suite: mean at the better learning rate. MQAR is teacher-forced answer-character accuracy, WikiText is character perplexity (lower is better), and all other rows are greedy generation exact (\%). Task definitions for all 3M-suite tasks are in our repository.}
\label{tab:laLTE}
\begin{tabular}{@{}lrr@{}}
\toprule
Task & laLTE & SSA \\
\midrule
Cued copy & 100.00 & 98.70 \\
Associative recall & 100.00 & 92.19 \\
Copy join & 100.00 & 100.00 \\
Alias chain (3 hops) & 89.84 & 48.70 \\
$k$-hop ($N=100$) & 19.01 & 19.53 \\
MQAR, 8 pairs & 99.77 & 99.18 \\
MQAR, 16 pairs & 99.20 & 72.23 \\
MQAR, 32 pairs & 90.79 & 30.44 \\
MQAR, 64 pairs & 97.98 & 24.22 \\
MKVT $(4,4)$ & 0.00 & 100.00 \\
MKVT $(4,8)$ & 0.78 & 99.61 \\
MKVT $(8,8)$ & 0.00 & 49.61 \\
MKVT $(4,16)$ & 1.56 & 2.34 \\
RULER variable tracking & 81.25 & 100.00 \\
Variable tracking & 3.91 & 8.20 \\
WikiText variable tracking & 17.19 & 1.56 \\
$n$-gram suffix (256) & 7.03 & 42.19 \\
WikiText (character PPL) & 2.57 & 2.74 \\
\bottomrule
\end{tabular}
\end{table}

At this scale, the full-attention Transformer with SSA's reader dimensions fails the same tasks: alias chains (39.58\%), $k$-hop (17.19\%), MQAR with 64 pairs (21.65\%), MKVT $(4,24)$ (0.78\%), and variable tracking (0.00\%), so these gaps reflect the small backbone rather than retention. The exceptions are MKVT $(8,8)$, and one seed each of MKVT $(4,16)$ and 32-pair MQAR, which the Transformer solves and SSA does not.
Table~\ref{tab:ssa_variants} compares SSA's membership and precision variants at equal state size in bytes, and
Table~\ref{tab:ssa_hybrids} hybrids with GDN layers. Layer or head-wise membership, and BF16 storage with 30
chunks, reach 94--100\% on MKVT $(8,8)$ and 16-pair MQAR, where shared FP32 membership with 14 chunks reaches
49.61\% and 72.23\%.

\begin{table}[h!]
\centering\small
\caption{SSA variants in the 3M suite: chunk membership shared across layers and heads (the SSA column of
Table~\ref{tab:laLTE}), independent per layer, or independent per layer and head, with FP32 keys and values and 14
retained chunks or BF16 and 30 chunks. Mean at the better learning rate; metrics as in Table~\ref{tab:laLTE}.
$^\dagger$One or more runs in the cell were stopped by the recovery liveness guard (no accepted candidate in 25
hard-phase epochs resulting in early termination) and are excluded. --: no completed run.}
\label{tab:ssa_variants}
\begin{tabular}{@{}lrrrrrr@{}}
\toprule
 & \multicolumn{3}{c}{FP32, 14 chunks} & \multicolumn{3}{c}{BF16, 30 chunks} \\
\cmidrule(lr){2-4}\cmidrule(l){5-7}
Task & Shared & Layer & Head & Shared & Layer & Head \\
\midrule
Cued copy & 98.70 & 99.22 & 99.22 & 97.40 & 98.18 & 100.00 \\
Associative recall & 92.19 & 95.05 & -- & 99.22 & 99.48 & -- \\
Copy join & 100.00 & 100.00 & 100.00 & 100.00 & 100.00 & 100.00 \\
Alias chain (3 hops) & 48.70 & 44.53 & 50.52 & 71.09 & 38.28 & 30.47 \\
$k$-hop ($N=100$) & 19.53 & 22.14 & 21.88 & 19.53 & 19.27 & 19.01 \\
MQAR, 8 pairs & 99.18 & 93.66 & 99.57 & 84.12 & 99.10 & 99.35 \\
MQAR, 16 pairs & 72.23 & 96.77 & 93.84 & 99.79 & 99.69 & 99.29$^\dagger$ \\
MQAR, 32 pairs & 30.44 & 30.71 & 30.33 & 30.74$^\dagger$ & 41.13 & 65.37 \\
MQAR, 64 pairs & 24.22 & 24.18 & 23.91 & 20.30 & 14.16$^\dagger$ & 19.88$^\dagger$ \\
MKVT $(4,4)$ & 100.00 & 100.00 & 99.61 & 99.61 & 99.61 & 99.61 \\
MKVT $(4,8)$ & 99.61 & 99.22 & 99.61 & 100.00 & 99.61 & 99.61 \\
MKVT $(8,8)$ & 49.61 & 99.61 & 99.22 & 94.92 & 100.00 & 98.44 \\
MKVT $(4,16)$ & 2.34 & 44.53 & 44.14 & 49.61 & 58.59 & 80.08 \\
RULER variable tracking & 100.00 & 100.00 & 100.00 & 100.00 & 100.00 & 99.74 \\
Variable tracking & 8.20 & 7.81 & 3.52 & 7.81 & 11.72 & 17.19 \\
WikiText variable tracking & 1.56 & 1.56 & 1.56 & 1.56 & 1.56 & 1.56 \\
$n$-gram suffix (256) & 42.19 & 42.97 & 73.44 & 0.00 & 0.00 & 0.00 \\
WikiText (character PPL) & 2.74 & 2.73 & 2.76 & 2.65 & 2.66 & 2.68 \\
\bottomrule
\end{tabular}
\end{table}

\begin{table}[h!]
\centering\small
\caption{Hybrids in the 3M suite: three GDN layers followed by one SSA layer, with one membership set for the SSA
layer or one per head, and the same stack with a full-attention fourth layer (growing memory). Mean at the better
learning rate; metrics and markers as in Table~\ref{tab:ssa_variants}.}
\label{tab:ssa_hybrids}
\begin{tabular}{@{}lrrrrr@{}}
\toprule
 & \multicolumn{2}{c}{SSA, FP32, 14 chunks} & \multicolumn{2}{c}{SSA, BF16, 30 chunks} & Full \\
\cmidrule(lr){2-3}\cmidrule(lr){4-5}
Task & Layer & Head & Layer & Head & attention \\
\midrule
Cued copy & 100.00 & 100.00 & 100.00 & 100.00 & 100.00 \\
Associative recall & 100.00 & 100.00 & 100.00 & 100.00 & 100.00 \\
Copy join & 100.00 & 100.00 & 100.00 & 100.00 & 100.00 \\
Alias chain (3 hops) & 94.01 & 90.89 & 85.16 & 72.14 & 88.54 \\
$k$-hop ($N=100$) & 13.54 & 16.41 & 13.28 & 14.32 & 11.46 \\
MQAR, 8 pairs & 99.95 & 99.94 & 99.91 & 99.98 & 99.98 \\
MQAR, 16 pairs & 99.88 & 99.94 & 99.94 & 100.00$^\dagger$ & 99.82 \\
MQAR, 32 pairs & 99.79 & 99.81$^\dagger$ & 99.80 & 99.96 & 99.77 \\
MQAR, 64 pairs & 73.71 & 59.33 & 99.93 & 99.92$^\dagger$ & 99.94 \\
MKVT $(4,4)$ & 61.33 & 62.50 & 99.22 & 99.22 & 48.44 \\
MKVT $(4,8)$ & 52.34 & 53.91 & 99.61 & 99.61 & 48.44 \\
MKVT $(8,8)$ & 57.03 & 52.73 & 55.08 & 55.47 & 48.83 \\
MKVT $(4,16)$ & 5.47 & 6.25 & 73.05 & 66.80 & 1.56 \\
RULER variable tracking & 100.00 & 100.00 & 100.00 & 100.00 & 100.00 \\
Variable tracking & 47.66 & 48.05 & 96.88$^\dagger$ & 3.91$^\dagger$ & 48.05 \\
WikiText variable tracking & 2.34 & 1.56 & 96.09 & 1.56 & 21.09 \\
$n$-gram suffix (256) & 92.97 & 80.47 & 50.00 & 61.72 & 97.66 \\
WikiText (character PPL) & 2.72 & 2.72 & 2.69 & 2.68 & 2.52 \\
\bottomrule
\end{tabular}
\end{table}

\section{Scorer architecture}
\label{app:scorer}

The scorer is a Transformer that runs once per eviction decision (Figure~\ref{fig:scorer}). The decision after chunk step $t$ involves the $M$ residents in slot order, the candidate (the chunk that has just left the window), and the $w$ window chunks.

\paragraph{Inputs.}
Let $e_{c,i}\in\mathbb R^{d}$ be the reader's token embedding of token $i$ of chunk $c$; the scorer does not read hidden states. With scorer width $d_s$, each token enters as
\begin{equation}
 x_{c,i}=W_{\mathrm{in}}e_{c,i}+\pi_i+u_{\sigma(c)},
\end{equation}
where $W_{\mathrm{in}}\in\mathbb R^{d_s\times d}$ has no bias, $\pi_i$ is a learned embedding of the position within the chunk ($i=1,\dots,C$), and $u_{\sigma(c)}$ is a learned embedding of the chunk's role: $\sigma(c)=j$ for the resident in slot $j$, $M+1$ for the candidate, and $M+2$ for every window chunk. Each resident and the candidate also has a summary token $z_c=z+u_{\sigma(c)}$, where $z$ is one learned vector. Window chunks have no summary token: they are context and are not scored. The input sequence is the $M+1$ summary tokens followed by the $(M+1+w)C$ tokens.

\paragraph{Blocks.}
The scorer has $n$ pre-norm blocks, $y\leftarrow y+\mathrm{Attn}(\mathrm{RMSNorm}(y))$ followed by $y\leftarrow y+\mathrm{SwiGLU}(\mathrm{RMSNorm}(y))$, with 4 attention heads, feed-forward width $4d_s$ and no attention mask: every position attends to every summary token and every token. In the last block only the summary tokens are queries. They attend to the whole sequence, and only they pass through the feed-forward layer, since no other output is used.

\paragraph{Scores and eviction.}
A linear head without bias maps each final summary token to a keep score $s_c$. At evaluation the chunk with the lowest of the $M+1$ scores is evicted; in hard training the evicted chunk is sampled with probability proportional to $\exp(-s_c/T_e)$, with sampling temperature $T_e$ (App.~\ref{app:post-training}). The candidate can be the evicted chunk, which leaves the residents unchanged.

\paragraph{Sharing and size.}
The scores are computed once per decision and shared by all layers and heads of the reader. With $n=2$ the scorer has $32d_s^2+d_s(d+M+C+8)$ parameters. In the 3M family $d_s=160$ (0.87M parameters); at 400M $d_s=640$ (13.8M parameters, 3.45\% of the model). The larger scorer of \S\ref{sec:scaling_posttraining} has four blocks of width 896 (52.4M parameters, 13\%).

\section{Hidden-state scorer ablation}
\label{appendix:hiddenSSA}
We compared the token-embedding scorer with scorers that read the reader's hidden states early on. A hidden-state scorer's inputs depend on memory contents, so parallel training needs fixed-point iteration over the membership mask, at two to three times the cost per step. We did not see a clear performance improvement, so we retired this version of the architecture. Code for these experiments is also available in the repository.

\section{400M pretraining and post-training details}
\label{app:post-training}

\subsection{Pretraining and model configurations}

All scaled models use seed 100 and are pretrained on 15B
SlimPajama content tokens, excluding inserted document-boundary
tokens from this budget. We use the Mistral-7B tokenizer~\citep{jiang2023mistral} and a nominal
global batch of 524,288 input tokens. Pretraining uses AdamW~\citep{loshchilov2019adamw} with
peak learning rate $4\times10^{-4}$, 1,000 warm-up updates, and
cosine decay to zero. AdamW uses
$(\beta_1,\beta_2)=(0.9,0.95)$, weight decay 0.01,
$\epsilon=10^{-8}$, and gradient clipping at one.
Training uses BF16 precision and reader dropout 0.1. These settings are already set in the provided code.

Transformer pretraining contexts are 576, 2,048, or 8,192 tokens.
GDN, HOLA, and all SSA variants are pretrained at 2,048 tokens.
SSA uses 16-token chunks, 34 retained chunks, and one recent-window
chunk, giving a maximum readable budget
$R=(34+1+1)\times16=576$ tokens.
SSA pretraining allocates the first 40\% of content tokens to the
soft phase and the remaining 60\% to the hard phase.
Recovery is disabled in pretraining except in the explicitly labeled
recovery-pretrained variant.

The original SSA scorer has two layers of width 640.
The larger-scorer variant has four layers of width 896,
containing 52,353,280 parameters.
Its reader feed-forward width is reduced from 3424 to 2904,
yielding 347,622,400 reader parameters and 399,975,680 total
parameters. This experiment reallocates parameters between reader
and scorer rather than enlarging the scorer while holding the
reader fixed, to keep the comparison parameter matched.

The post-training results in Table~\ref{tab:posttraining_final}
use the original SSA pretrained without recovery.
The recovery-pretrained variant is included in the pretraining
comparison.
We additionally evaluate the larger-scorer variant after
mixed-task post-training in
Appendix~\ref{app:posttraining_scorer_capacity}.
Those results are measured before dedicated $k$-hop specialization.
\paragraph{Pretraining evaluation windows.}
The pretraining language evaluations use the same reserved
SlimPajama test corpus before any task post-training.
Transformer-$576$ retains its fixed 576-token attention window.
The original Transformer-(2K) and Transformer-(8K) evaluations
permit full attention over the stated evaluation segment.
Consequently, the Transformer-(2K) result at 8K measures
unrestricted context extrapolation. All SSA evaluations use hard retention
with $R=576$.

\subsection{Persistent memory budgets}

Table~\ref{tab:400m_state} reports analytical persistent tensor
payloads for batch size one. These quantities exclude model weights,
optimizer state, training activations, temporary buffers, and
allocator overhead: we intend them as a common measure for comparing the memory contents of architectures that store information in different ways.
Transformer windows are enforced per layer during task training
and evaluation.

\begin{table}[h!]
\centering
\small
\caption{Analytical persistent state at 400M scale, batch size one.
MiB denotes $2^{20}$ bytes.}
\label{tab:400m_state}
\begin{tabular}{@{}lr@{}}
\toprule
Model or attention window & Persistent state (MiB) \\
\midrule
Transformer, 576-token window & 54.000 \\
Transformer, 2K window       & 192.000 \\
Transformer, 8K window       & 768.000 \\
SSA, $R=576$                & 54.005 \\
GDN                         & 54.797 \\
HOLA                        & 54.680 \\
\bottomrule
\end{tabular}
\end{table}

Thus, SSA and Transformer-$576$ match readable-token budgets,
while SSA, GDN, and HOLA have approximately matched persistent
state sizes.

\subsection{Post-training stages and objectives}

The reported comparison consists of a joint pilot, mixed-task
refinement, and dedicated $k$-hop specialization
(Table~\ref{tab:400m_training_budget}).
All stages use global batch size 64 and seed 100.
The models receive the same primary training examples and stored
example order within each stage.
The refinement uses fresh examples rather than repeating the
pilot corpus.

\begin{table}[h!]
\centering
\small
\caption{Training budgets for the final 400M comparison.
Learning rates are reader peak rates.}
\label{tab:400m_training_budget}
\begin{tabular}{@{}lrrr@{}}
\toprule
Stage & Updates & Examples & Reader LR \\
\midrule
Joint pilot       & 1,560 & 99,840  & $10^{-4}$ \\
Mixed refinement  & 6,240 & 399,360 & $10^{-5}$ \\
$k$-hop specialization & 1,760 & 112,640 & $3\times10^{-5}$ \\
\midrule
Total             & 9,560 & 611,840 & --- \\
\bottomrule
\end{tabular}
\end{table}

The pilot and refinement use fresh AdamW optimizers, with
156 and 624 warm-up updates, respectively, followed by cosine
decay to zero. Dedicated $k$-hop training preserves the refinement's
AdamW moments, warms up for 20 updates, and holds the reader
learning rate at $3\times10^{-5}$.
Its successive 440, 440, and 880-update tranches use fresh data
without additional optimizer resets or warm-up periods. We chose this setup because we observed that $k$-hop needs especially large amounts of data.
The trainable SSA scorer uses peak learning rate $10^{-4}$. During the soft phase, $\lambda$ rises linearly to one and $\tau$ decays geometrically from one to $\tau_{\min}=0.05$, both as functions of progress through the soft phase. Optimizer betas, weight decay, clipping, precision, and reader
dropout follow the settings above, and the exact details are available in the repository.

\paragraph{Task mixture and loss.}
The pilot and refinement allocate equal example weight to MKVT,
atomic MQAR, $k$-hop, and RULER. RULER's quarter of the mixture is
divided equally among its 13 tasks, which may explain its low scores in all models.
Losses are normalized per example so that longer answers do not
automatically receive greater weight.
MKVT and RULER examples use separately averaged answer and
non-answer language cross-entropies, with unit weight on each.
Atomic MQAR and $k$-hop use answer loss only.

\paragraph{SSA training phases and recovery.}
The pilot uses 624 soft updates followed by 936 hard updates.
Refinement uses 624 soft updates followed by 5,616 hard updates.
Attention distillation has weight one during the pilot's soft
phase and is disabled during refinement, as ablations showed that it slightly reduced performance.

Attention distillation trains the scorer to match the reader's own soft-phase read: a no-gradient forward pass measures how much attention the reader's supervised positions place on each candidate chunk under the current soft retention weights, and the scorer's softmax over those candidates is fit to this distribution with a KL loss (temperature one). Examples that place no attention on any candidate are skipped.

The original pilot selects recovery examples globally by answer
loss. The reported refinement instead reserves two recovery slots
per family per batch of 64, with RULER reservations rotated across
its tasks, as certain tasks were found to dominate recovery allocations and to starve the scorer of data. Unused reservations are not transferred to another
family. This is \textit{different} from equal weighting of the primary task
losses, which applies in both stages.
Hard training samples eviction decisions; evaluation uses
deterministic lowest-score eviction. Each eviction is sampled with probability proportional to $\exp(-s_c/T_e)$ over candidate scores $s_c$, where the sampling temperature $T_e$ starts at one and is multiplied by 0.99 every 128 hard updates. Evaluation is the $T_e\to0$ limit.

\paragraph{Frozen-scorer specialization.}
Both SSA specialization variants start from the same checkpoint
after 7,800 mixed-task updates.
Dedicated $k$-hop training begins with five soft updates and then
uses hard training. This is the default training recipe, and we do not expect it to make a difference here, as all data fits within the context.

In the frozen condition, scorer parameters and their optimizer
states remain unchanged throughout all 1,760 specialization
updates. Scorer inputs are detached, and the scorer-to-reader
straight-through gradient path, recovery updates, and attention
distillation are disabled.
\subsection{Benchmarks and evaluation metrics}

\paragraph{MKVT.}
At each of 2K, 4K, and 8K context, we evaluate six settings:
$(K,B)\in\{(4,4),(4,8),(4,16),(4,24),(8,8),(16,16)\}$,
where $K$ is the number of query-eligible keys and $B$ the total
number of bindings. Each setting has 500 test examples.
The reported value is the mean generated whole-answer exact
accuracy across the six settings at that context.

\paragraph{Atomic MQAR.}
We evaluate eight sequence-length/pair-count settings:
$(64,4)$, $(128,8)$, $(256,16)$, $(512,64)$,
$(1024,64)$, $(1024,128)$, $(2048,128)$, and $(4096,128)$.
Each setting has 1,000 test examples.
We report the mean query accuracy across these settings.

\paragraph{RULER.}
We evaluate all 13 RULER tasks at 4K and 8K, using 500 test
examples per task and context.
The reported score is the unweighted mean of the task-specific
scores across all 13 tasks.

\paragraph{Higher-hop specialization.}
The dedicated task uses $N=256$ source positions and
$h\in\{4,8,16\}$, with balanced training exposure
across hop counts.
Task symbols map to individual tokens in the
vocabulary.
The fixed test set contains 256 examples per hop count.
Each example has 256 answer tokens.

The main text reports generated-answer token accuracy excluding the null symbol, which makes up 5\%, 10\% and 19\% of the answer tokens at 4, 8 and 16 hops. Whole-sequence exact match, for which every generated answer token must match the reference, is unstable for answers of this length: a model at 99.5\% token accuracy produces roughly half of its 16-hop answers entirely correctly, and small differences between runs move it by many points. Table~\ref{tab:khop_exact} reports it together with token accuracy over all answer tokens. All three metrics come from the same predictions.

\begin{table}[h!]
\centering\small
\caption{400M $k$-hop ($N=256$) after specialization, by hop count: greedy whole-sequence exact match / token accuracy over all answer tokens (\%). The main text reports token accuracy excluding the null symbol (Table~\ref{tab:posttraining_final}).}
\label{tab:khop_exact}
\begin{tabular}{@{}lccc@{}}
\toprule
Model & $h=4$ & $h=8$ & $h=16$ \\
\midrule
SSA, frozen scorer & 96.48 / 99.985 & 94.92 / 99.971 & 65.63 / 99.667 \\
SSA, trainable scorer & 97.66 / 99.985 & 95.31 / 99.974 & 56.25 / 99.600 \\
Transformer $576\rightarrow576$ & 98.83 / 99.995 & 92.58 / 99.969 & 55.47 / 99.513 \\
Transformer $2\mathrm{K}\rightarrow576$ & 96.48 / 99.983 & 65.23 / 99.748 & 0.00 / 57.008 \\
Transformer $2\mathrm{K}\rightarrow2\mathrm{K}$ & 78.52 / 99.768 & 0.00 / 39.908 & 0.00 / 38.676 \\
Transformer $2\mathrm{K}\rightarrow8\mathrm{K}$ & 5.86 / 83.319 & 0.00 / 35.718 & 0.00 / 38.831 \\
Transformer $8\mathrm{K}\rightarrow8\mathrm{K}$ & 99.22 / 99.997 & 95.31 / 99.979 & 47.27 / 99.545 \\
GDN & 0.00 / 29.486 & 0.00 / 32.402 & 0.00 / 38.789 \\
HOLA & 0.00 / 30.264 & 0.00 / 32.259 & 0.00 / 38.562 \\
\bottomrule
\end{tabular}
\end{table}

The 20M comparison is teacher-forced and the 400M comparison is greedy, and
their task distributions differ.
The smaller-scale task uses $N=100$, 100-token answers, and mixed
hop counts from zero through sixteen.
The 400M task uses longer answers and separately evaluated hop
counts.

The unpadded prompt-plus-answer sequences contain 517 tokens,
so the complete useful sequence fits within the 576-token reader
budget. Training tensors are right-padded to 8K, after all supervised
positions. This task therefore tests multi-hop computation without
requiring SSA to evict useful information.

\paragraph{Decoding and language evaluation.}
Generation is greedy. MKVT and $k$-hop generation use the reference
answer length; RULER uses its task-specific generation limits.
SSA uses hard retention at evaluation.
Task scores are reported as percentages.
Language evaluation uses the same reserved SlimPajama corpus,
containing approximately 1.05M tokens, segmented at 2K or 8K.
We report NLL per predicted token in the pretraining table and
its exponential, perplexity, in the post-training table, as is common for pretraining and post-training results.
Lower values indicate better language modeling in both.
All 400M results use one seed, although ablations over several settings were run, and we have run the same seeds multiple times.

\subsection{Scorer capacity and recovery allocation}
\label{app:posttraining_scorer_capacity}

We compare the original two-layer, width-640 scorer with the
four-layer, width-896 scorer described above.
The larger scorer is paired with a narrower reader feed-forward
network, preserving approximately 400M total parameters and the
same $R=576$ readable-token budget.

\paragraph{Recovery allocation and run-to-run variation.}
Mixed-task refinement reserves recovery slots per task family: MKVT, MQAR, $k$-hop, and RULER
each receive two of the eight recovery slots per batch of 64 examples, with RULER's share
balanced across its 13 tasks, and unused slots are not transferred between families.
We first tried this allocation in a 1,560-update pilot, with the same architecture, pretrained
checkpoint, 99,840 primary training examples, schedule and learning rates as a pilot that
selects recovery examples globally by loss.

The pilot with per-family slots reaches 89.70\%, 91.67\%, and 90.90\% MKVT accuracy at 2K, 4K,
and 8K, against 32.63\%, 17.77\%, and 8.20\% with global selection
(Table~\ref{tab:equal_recovery_pilot}). We do not attribute this gap to the allocation.
Two reruns of the per-family pilot with the same configuration and seed reach 29.7\% and
28.8\% at 2K. At this budget a run either learns to match the query key or stays near 30\%,
and the same recipe can end on either side (Table~\ref{tab:ablations}). With the full 7,800-update budget the model
reaches 98.40\% on 8K MKVT (Table~\ref{tab:scorer_capacity_refinement}).

\begin{table}[h!]
\centering
\small
\caption{Original-size SSA after 1,560 post-training updates
on the same 99,840 primary examples, with one seed.
Accuracy metrics are percentages; lower perplexity is better.
The global-recovery column is historical and also differs in
implementation version and GPU/microbatch execution.
Two same-seed reruns of the equal-family pilot reached 29.7 and 28.8 on MKVT at 2K (see text).}
\label{tab:equal_recovery_pilot}
\begin{tabular}{@{}lrr@{}}
\toprule
Metric & Global recovery & Equal-family recovery \\
\midrule
MKVT, 2K       & 32.63 & 89.70 \\
MKVT, 4K       & 17.77 & 91.67 \\
MKVT, 8K       &  8.20 & 90.90 \\
MQAR           & 99.95 & 99.95 \\
RULER, 4K      & 21.86 & 56.65 \\
RULER, 8K      & 12.64 & 45.36 \\
PPL, 2K $\downarrow$ & 20.76 & 22.54 \\
PPL, 8K $\downarrow$ & 26.36 & 26.05 \\
\bottomrule
\end{tabular}
\end{table}

\paragraph{Allocating more capacity to the scorer.}
We also post-train the larger-scorer model, which reallocates
parameters from the reader's feed-forward layers to a four-layer,
width-896 scorer while keeping the total near 400M.
With equal-family recovery in both 1,560-update pilots, the
larger variant improves 8K MKVT from 90.90\% to 97.87\% and
8K RULER from 45.36\% to 55.59\%.
Both are single runs in the regime described above, so we do not read this gap as an
effect of scorer size.

After 7,800 mixed-task updates, the larger variant reaches 99.87\% on 8K MKVT, compared with 98.40\% for the original scorer, a small difference. This comparison is not fully matched, because the original model's pilot used global recovery (Table~\ref{tab:equal_recovery_pilot}), but this may not have a major effect. Their final RULER means are also nearly identical: 68.84\% versus 68.73\% at 4K and 61.87\% versus 61.90\% at 8K. The larger variant, however, has worse language perplexity, 22.05 versus 21.40 at 2K and 25.99 versus 24.56 at 8K. Thus, allocating more capacity to the scorer may improve early retrieval performance, but does not produce a corresponding improvement in final aggregate RULER performance and increases language perplexity. The complete comparisons follow.
\paragraph{Matched equal-recovery pilots.}
To separate the architecture comparison from recovery allocation,
we train both variants with equal-family recovery from the first
hard-phase update.
Both pilots use the same 99,840 primary examples and stored order,
global batch size 64, seed 100, 1,560-update schedule, and peak
reader and scorer learning rates of $10^{-4}$.
The first 624 updates use soft training with attention
distillation; the remaining 936 use hard training with recovery.
Equal recovery, as before, reserves two of eight slots for each task family,
with RULER's share balanced across its 13 tasks.
Unused reservations are not transferred between families.

Table~\ref{tab:scorer_capacity_pilot} reports these matched
pilots. The larger variant improves MKVT at all three contexts
and RULER at both contexts. MQAR is already near saturation.
Its 2K language perplexity is slightly lower, while its
8K perplexity is higher.

The mixed-task $k$-hop entries in the tables below average
greedy whole-sequence exact match across the original
$N\in\{100,256,512,1024\}$ evaluation settings, with 500 examples
per setting. These are evaluations \textbf{before the dedicated
higher-hop specialization} and are distinct from the
$N=256$, $h\in\{4,8,16\}$ results in
Table~\ref{tab:posttraining_final}.

\begin{table}[h!]
\centering
\small
\caption{
Matched SSA pilots after 1,560 updates. Both use equal-family recovery.
Accuracy metrics are percentages; lower perplexity is better.
}
\label{tab:scorer_capacity_pilot}
\begin{tabular}{@{}lrr@{}}
\toprule
Metric & Original scorer & Larger scorer \\
\midrule
MKVT, 2K                  & 89.70  & 98.13 \\
MKVT, 4K                  & 91.67  & 98.40 \\
MKVT, 8K                  & 90.90  & 97.87 \\
MQAR                      & 99.951 & 99.959 \\
RULER, 4K                 & 56.65  & 63.08 \\
RULER, 8K                 & 45.36  & 55.59 \\
Mixed-task $k$-hop exact   & 4.05   & 3.80 \\
PPL, 2K $\downarrow$      & 22.54  & 22.46 \\
PPL, 8K $\downarrow$      & 26.05  & 27.10 \\
\bottomrule
\end{tabular}
\end{table}

\paragraph{Mixed-task refinement.}
We additionally compare the larger variant with the existing
original-scorer refinement at 3,120 and 7,800 cumulative updates,
corresponding to 199,680 and 499,200 primary examples.
The models receive the same primary refinement examples and
use the same global batch and learning-rate schedules.
Both refinements use equal-family recovery and disable
attention distillation.

At 3,120 updates, the larger variant has higher MKVT and RULER
scores. By 7,800 updates, the original model nearly matches
its aggregate RULER performance, while the larger variant
retains a 1.40--1.67 percentage-point advantage on MKVT.
The larger variant has worse final language perplexity at
both evaluation contexts.

\begin{table}[h!]
\centering
\small
\setlength{\tabcolsep}{5pt}
\caption{
SSA mixed-task refinement at matched cumulative update counts,
before dedicated $k$-hop specialization. Both use equal-family recovery during refinement, though they differ in post-training pilot recovery settings (see the repository for more details).
}
\label{tab:scorer_capacity_refinement}
\begin{tabular}{@{}lrrrr@{}}
\toprule
& \multicolumn{2}{c}{3,120 total updates}
& \multicolumn{2}{c}{7,800 total updates} \\
\cmidrule(lr){2-3}
\cmidrule(lr){4-5}
Metric & Original & Larger & Original & Larger \\
\midrule
MKVT, 2K
& 92.03 & 99.50 & 98.43 & 99.83 \\
MKVT, 4K
& 92.57 & 99.33 & 98.23 & 99.90 \\
MKVT, 8K
& 92.63 & 98.83 & 98.40 & 99.87 \\
MQAR
& 99.957 & 99.977 & 99.963 & 99.995 \\
RULER, 4K
& 58.13 & 66.08 & 68.73 & 68.84 \\
RULER, 8K
& 47.55 & 58.71 & 61.90 & 61.87 \\
Mixed-task $k$-hop exact
& 4.05 & 3.85 & 4.05 & 4.20 \\
PPL, 2K $\downarrow$
& 21.97 & 21.89 & 21.40 & 22.05 \\
PPL, 8K $\downarrow$
& 23.77 & 24.33 & 24.56 & 25.99 \\
\bottomrule
\end{tabular}
\end{table}

\paragraph{RULER subtask differences.}
The nearly unchanged final RULER means conceal differences
between tasks.
At 8K, the larger variant improves NIAH multi-key-1 from
90.60\% to 97.20\% and multi-query retrieval from 84.60\%
to 92.10\%.
However, CWE decreases from 53.42\% to 34.80\%.
Table~\ref{tab:scorer_capacity_ruler} reports all 13 tasks.

\begin{table}[h!]
\centering
\small
\setlength{\tabcolsep}{5pt}
\caption{
RULER task scores (\%) after 7,800 mixed-task updates.
Each task and context uses 500 test examples.
}
\label{tab:scorer_capacity_ruler}
\begin{tabular}{@{}lrrrr@{}}
\toprule
& \multicolumn{2}{c}{4K}
& \multicolumn{2}{c}{8K} \\
\cmidrule(lr){2-3}
\cmidrule(lr){4-5}
Task & Original & Larger & Original & Larger \\
\midrule
NIAH single-1    & 100.00 & 100.00 & 99.80 & 100.00 \\
NIAH single-2    & 100.00 & 100.00 & 99.40 & 99.80 \\
NIAH single-3    & 96.40  & 97.80  & 88.80 & 88.40 \\
NIAH multi-key-1 & 99.40  & 99.80  & 90.60 & 97.20 \\
NIAH multi-key-2 & 15.60  & 15.00  & 8.80  & 10.80 \\
NIAH multi-key-3 & 5.00   & 4.80   & 1.40  & 1.20 \\
NIAH multi-value & 92.75  & 95.35  & 91.35 & 91.65 \\
NIAH multi-query & 93.10  & 99.15  & 84.60 & 92.10 \\
VT              & 98.20  & 98.88  & 94.96 & 96.32 \\
CWE             & 78.76  & 69.28  & 53.42 & 34.80 \\
FWE             & 91.27  & 91.67  & 74.80 & 74.20 \\
QA-1            & 6.00   & 7.20   & 4.20  & 4.80 \\
QA-2            & 17.00  & 16.00  & 12.60 & 13.00 \\
\midrule
Mean            & 68.73  & 68.84  & 61.90 & 61.87 \\
\bottomrule
\end{tabular}
\end{table}

\subsection{Forgetting during $k$-hop specialization}
\label{app:posttraining_forgetting}

We measure forgetting relative to each model's checkpoint after
7,800 mixed-task updates, immediately before dedicated $k$-hop
training.
Changes therefore span all 1,760 specialization updates.
Accuracy changes are final minus initial scores in percentage
points; negative values indicate degradation.
Positive perplexity changes indicate degradation.

Freezing the SSA scorer reduces MKVT forgetting:
the 8K exact score falls by 0.87 percentage points, compared
with 1.83 points when the scorer remains trainable.
The corresponding RULER declines are 5.61 and 5.50 points,
and perplexity increases are 2.44 and 2.35.
Thus, freezing improves MKVT retention in this seed without
consistently reducing forgetting across the other metrics.

In general, SSA's forgetting is on a similar scale to that of the other architectures, and we do not see it as a concern.

\begin{table*}[h!]
\centering
\small
\caption{
Task forgetting over the full 1,760-update $k$-hop specialization.
Values are changes in percentage points;
negative values indicate degradation.
}
\label{tab:posttraining_task_forgetting}
\setlength{\tabcolsep}{5pt}
\resizebox{\textwidth}{!}{%
\begin{tabular}{lrrrrrr}
\toprule
& \multicolumn{3}{c}{$\Delta$ MKVT}
& $\Delta$ MQAR
& \multicolumn{2}{c}{$\Delta$ RULER} \\
\cmidrule(lr){2-4}
\cmidrule(lr){6-7}
Model & 2K & 4K & 8K & Query & 4K & 8K \\
\midrule
SSA, frozen scorer
& $-0.33$ & $-0.10$ & $-0.87$ & $-0.009$ & $-4.59$ & $-5.61$ \\
SSA, trainable scorer
& $-2.30$ & $-2.27$ & $-1.83$ & $-0.011$ & $-4.73$ & $-5.50$ \\
\midrule
Transformer $576\rightarrow576$
& $-0.30$ & $+0.03$ & $0.00$ & $-0.029$
& $-2.27$ & $-1.00$ \\
Transformer $2\mathrm{K}\rightarrow576$
& $+0.17$ & $-0.10$ & $-0.07$ & $-0.306$ & $-3.82$ & $-1.72$ \\
Transformer $2\mathrm{K}\rightarrow2\mathrm{K}$
& $-0.03$ & $-0.63$ & $-4.10$ & $-0.013$ & $-6.17$ & $-9.45$ \\
Transformer $2\mathrm{K}\rightarrow8\mathrm{K}$
& $-0.03$ & $-0.03$ & $-0.20$ & $-0.003$ & $-9.21$ & $-11.52$ \\
Transformer $8\mathrm{K}\rightarrow8\mathrm{K}$
& $0.00$ & $0.00$ & $-0.07$ & $-0.002$ & $-11.89$ & $-9.30$ \\
\midrule
GDN
& $-0.07$ & $-0.27$ & $-0.33$ & $-0.009$ & $-3.40$ & $-2.61$ \\
HOLA
& $-0.60$ & $-0.40$ & $-0.23$ & $-0.095$ & $-9.55$ & $-11.57$ \\
\bottomrule
\end{tabular}%
}
\end{table*}

\begin{table}[h!]
\centering
\small
\caption{
Change in held-out language perplexity over the full
1,760-update specialization.
Positive values indicate degradation.
}
\label{tab:posttraining_language_forgetting}
\begin{tabular}{@{}lrr@{}}
\toprule
Model & $\Delta$ PPL, 2K & $\Delta$ PPL, 8K \\
\midrule
SSA, frozen scorer    & $+1.59$ & $+2.44$ \\
SSA, trainable scorer & $+1.67$ & $+2.35$ \\
\midrule
Transformer $576\rightarrow576$
& $+1.43$ & $+1.40$ \\
Transformer $2\mathrm{K}\rightarrow576$
& $+2.21$ & $+2.15$ \\
Transformer $2\mathrm{K}\rightarrow2\mathrm{K}$
& $+1.59$ & $+1.61$ \\
Transformer $2\mathrm{K}\rightarrow8\mathrm{K}$
& $+1.56$ & $+2.89$ \\
Transformer $8\mathrm{K}\rightarrow8\mathrm{K}$
& $+2.69$ & $+2.18$ \\
\midrule
GDN  & $+1.95$ & $+1.98$ \\
HOLA & $+2.34$ & $+2.45$ \\
\bottomrule
\end{tabular}
\end{table}

\paragraph{Learning-rate sensitivity of the 2K-pretrained Transformers.}
Transformer 2K$\rightarrow$2K and 2K$\rightarrow$8K both perform clearly worse than 2K$\rightarrow$576. Our best explanation for this is training dynamics, though this is uncertain. These models saturate $k$-hop at higher learning rates, so this is not a capability gap; this is expected, as their model code is the same as that of all the other Transformers. However, the runs with a higher learning rate had substantially worse perplexity, around 29 instead of 19--22. We keep the shared schedule (Table~\ref{tab:400m_training_budget}) for all models because it gave the best results overall: the higher learning rate improved $k$-hop for these two models but caused more forgetting on the other tasks and in held-out perplexity.

\paragraph{Intermediate $k$-hop performance.}
Table~\ref{tab:400m_khop_progress} reports generated-answer
token accuracy at matched intermediate checkpoints.
At the recorded 16-hop readouts, all four models first exceed
99\% token accuracy at 1,760 updates.

\begin{table}[h!]
\centering
\small
\caption{
Generated 16-hop token accuracy (\%), excluding the null symbol, at matched numbers
of dedicated $k$-hop updates.
}
\label{tab:400m_khop_progress}
\begin{tabular}{@{}lrrr@{}}
\toprule
Model & 880 & 1,320 & 1,760 \\
\midrule
SSA, frozen scorer
& 26.16 & 49.52 & 99.61 \\
SSA, trainable scorer
& 25.96 & 34.79 & 99.54 \\
Transformer $576\rightarrow576$
& 25.97 & 47.04 & 99.47 \\
Transformer $8\mathrm{K}\rightarrow8\mathrm{K}$
& 26.29 & 56.81 & 99.46 \\
\bottomrule
\end{tabular}
\end{table}

\subsection{Component ablations}
\label{app:ablations}

Each run starts from the same pretrained SSA (no recovery during pretraining) and trains for 1,560 updates on the mixed tasks, 624 soft and then 936 hard, with the per-family recovery slots of Appendix~\ref{app:posttraining_scorer_capacity}. Each arm changes one component (Table~\ref{tab:ablations}). Runs A and B of an arm share their configuration and seed (100), and differ only through GPU nondeterminism. The third run of the full recipe is the equal-family pilot of Table~\ref{tab:equal_recovery_pilot}.

\begin{table}[h!]
\centering\small
\caption{Component ablations during 400M post-training at the pilot budget. MKVT exact match (\%) at 2K / 4K / 8K. ``Learned'' counts the runs that left the plateau near 30\% at 2K. $^{\ast}$The third run is the original pilot (89.7 / 91.7 / 90.9). $^{\dagger}$Mean over four of the six 2K settings.}
\label{tab:ablations}
\setlength{\tabcolsep}{4pt}
\resizebox{\linewidth}{!}{%
\begin{tabular}{@{}llccc@{}}
\toprule
Arm & Change from the full recipe & Run A & Run B & Learned \\
\midrule
Full recipe & none & 29.7 / 15.4 / 7.8 & 28.8 / 14.7 / 7.3 & 1 of 3$^{\ast}$ \\
No recovery & counterfactual recovery off & 28.5 / 14.8 / 8.0 & 29.3 / 14.4 / 7.1 & 0 of 2 \\
No STE & straight-through read gradient off & 29.1 / 15.9 / 8.0 & 54.0 / 41.9 / 31.5 & 1 of 2, in part \\
Hard only & no soft phase: 1,560 hard updates & 79.2 / 69.8 / 64.4 & 97.6$^{\dagger}$ / 97.4 / 95.7 & 2 of 2 \\
Contextual, no rescoring & each chunk scored once, on entry & 84.4 / 83.3 / 80.0 & 99.3 / 99.1 / 98.4 & 2 of 2 \\
Non-contextual & the scorer sees only the new chunk & 99.9 / 99.8 / 98.3 & 99.8 / 99.4 / 98.0 & 2 of 2 \\
\bottomrule
\end{tabular}}
\end{table}

The hard-only arm replaces the 624 soft updates with hard updates, so it has 1,560 hard updates and 195 recovery passes, against 936 and 117 in the other arms. The two arms without rescoring change what the scorer sees. In the contextual arm it still sees the residents, the candidate and the window, but each chunk is scored once, when it becomes the candidate, and keeps that score. In the non-contextual arm it sees only the new chunk. MQAR is between 99.93\% and 99.97\% in every arm, and mixed-task $k$-hop does not differ between arms.

The outcome is bimodal: a run either learns to match the query key or stays near 30\% at 2K. With two or three runs per arm the arms cannot be ranked. The table shows that the full recipe is not more reliable than the simpler arms at this budget, and that recovery is the one component whose removal left both runs on the plateau. With the full budget of 7,800 updates the full recipe reaches 98.40\% on MKVT 8K (Table~\ref{tab:scorer_capacity_refinement}).

\subsection{Cache eviction}
\label{app:eviction}

\paragraph{Training free cache eviction.} Each method runs on the post-trained full-attention Transformers of Table~\ref{tab:posttraining_final}, the one pretrained at 2K and the one pretrained at 8K (Table~\ref{tab:eviction}). At SSA's budget the cache holds 34 chunks of 16 tokens, one window chunk and the current chunk (576 tokens), and attention scores are summed over heads and layers. The paper-setting rows follow each paper's own rule and cache size. SnapKV and paper-setting H$_2$O read the whole prompt with full attention, question included, before they select; MQAR has no prefill boundary in its teacher-forced format, so they were not run on it. The $k$-hop rows (517 tokens) fit every 576-token cache, so these methods equal full attention there and are left out.

\begin{table}[h!]
\centering\small
\caption{Training-free cache eviction on the post-trained 400M Transformers, at 8K (\%). Each cell gives the 2K-pretrained / the 8K-pretrained Transformer. SSA is its own model (trainable scorer, Table~\ref{tab:posttraining_final}).}
\label{tab:eviction}
\begin{tabular}{@{}lrccc@{}}
\toprule
Method & Cache (tokens) & MKVT 8K & RULER 8K & MQAR \\
\midrule
Full attention, no eviction & 8,192 & 99.6 / 99.9 & 70.7 / 76.9 & 99.99 / 99.98 \\
StreamingLLM \citep{xiao2024streamingllm} & 576 & 7.6 / 7.8 & 11.6 / 15.7 & 85.8 / 85.7 \\
H$_2$O \citep{zhang2023h2o} & 576 & 5.3 / 5.0 & 3.8 / 4.2 & 100.0 / 100.0 \\
TOVA \citep{oren2024tova} & 576 & 6.2 / 6.7 & 11.3 / 16.9 & 98.8 / 96.9 \\
SnapKV \citep{li2024snapkv}, sees the question & 576 & 99.1 / 81.4 & 51.0 / 50.0 & not run \\
H$_2$O, paper settings (20\% of the prompt) & about 1,600 & 40.3 / 19.6 & 24.1 / 21.3 & not run \\
TOVA, paper settings (1/8 of the context) & 1,024 & 6.4 / 11.3 & 6.8 / 13.6 & 98.3 / 96.9 \\
\midrule
SSA & 576 & 96.6 & 56.4 & 99.95 \\
\bottomrule
\end{tabular}
\end{table}

\paragraph{Learned eviction: Dynamic Memory Sparsification.}
DMS \citep{nawrot2025dms} starts from the 2K-pretrained 400M Transformer and follows the recorded post-training chain of the full-attention $2\mathrm{K}\rightarrow8\mathrm{K}$ Transformer: 9,560 updates with the same data, order, objective, optimizer and schedules, with the DMS gates and loss added and training rows trimmed of padding. Settings follow the paper and the official code: one eviction logit per layer and head from a zero-initialized projection of the normalized residual, Gumbel-sigmoid with $\tau=0.1$ and straight-through hard decisions, delayed eviction with a 256-token window, and a compression ratio that ramps by 1 every 100 updates to 24.87, which keeps 576 tokens at 8K. We did not search hyperparameters ourselves. At evaluation a streaming decoder applies DMS's own rule plus a cap of 576 readable tokens per head, after which DMS keeps 344 tokens per head at the end of 8K MKVT prompts. Each of the 96 layer--head pairs chooses separately, so the union of positions readable somewhere in the model is about 4,480 (59\% of the context), against 576 for SSA. Results use one seed (Table~\ref{tab:dms}).

\paragraph{Learned eviction: Dynamic Context Pruning.}
DCP \citep{anagnostidis2023dynamic} is trained the same way as DMS: from the 2K-pretrained Transformer through the same 9,560 updates, with training rows trimmed of padding. Each layer gets a learned interaction query and key of width 64 and a scalar bias, whose $\alpha$-sigmoid gates let later tokens drop earlier ones; drops are irreversible and shared by the layer's heads, and a regularizer weighted by $\gamma$ sets the sparsity. As in the paper, $\alpha$ rises from 1 to 8 on a cosine schedule and the gates become a step function at inference. We chose $\gamma$ with a rule fixed before training: among 0.3, 1 and 3, the value whose 390-update pilot kept closest to 576 tokens at the end of 8K training rows, which was $\gamma=1$ (521 tokens; also the paper's value). At evaluation a streaming decoder applies DCP's own rule plus a cap of 576 tokens per layer, removing the token whose running interaction logit is smallest when the cap binds. The fully trained model prunes far more than its pilot did: at the end of 8K MKVT prompts each layer keeps 62 tokens on average, and the union over its 24 layers is about 360 distinct positions. It is far worse than DMS and SSA on every task and in perplexity (Table~\ref{tab:dms}). Results use one seed.

\begin{table}[h!]
\centering\small
\caption{Dynamic Memory Sparsification and Dynamic Context Pruning against SSA and the full-attention Transformer they start from, after the same 9,560 post-training updates (\%, except perplexity). $k$-hop is $N=256$; its token accuracy excludes the null symbol. DCP decides per layer, so its tokens per head are its tokens per layer.}
\label{tab:dms}
\setlength{\tabcolsep}{3pt}
\resizebox{\linewidth}{!}{%
\begin{tabular}{@{}lcccc@{}}
\toprule
 & DMS & DCP & SSA & Transformer $2\mathrm{K}\rightarrow8\mathrm{K}$ \\
\midrule
Tokens readable per head at 8K & 344 & 62 & 576 (shared) & 8,192 \\
MKVT 2K / 4K / 8K & 99.93 / 99.03 / 95.43 & 23.93 / 17.90 / 8.63 & 96.13 / 95.97 / 96.57 & 99.97 / 99.97 / 99.60 \\
MQAR & 99.97 & 50.43 & 99.95 & 99.99 \\
RULER 4K / 8K & 63.77 / 50.84 & 12.62 / 13.46 & 64.00 / 56.40 & 78.27 / 70.72 \\
$k$-hop exact, $h=4$ / 8 / 16 & 87.50 / 1.95 / 0.00 & 0.00 / 0.00 / 0.00 & 97.66 / 95.31 / 56.25 & 5.86 / 0.00 / 0.00 \\
$k$-hop token, $h=4$ / 8 / 16 & 99.95 / 97.55 / 38.61 & 24.73 / 24.24 / 24.75 & 99.98 / 99.97 / 99.54 & 82.53 / 29.42 / 26.07 \\
Perplexity 2K / 8K & 20.92 / 22.21 & 56.79 / 48.07 & 23.07 / 26.91 & 19.73 / 22.28 \\
\bottomrule
\end{tabular}}
\end{table}

\subsection{$k$-hop induction inside 8K inputs}
\label{app:haystack}

We train at 8K context. The 258-token $k$-hop source block sits in fresh SlimPajama text, followed by the question and a 256-token answer (4, 8 or 16 hops), with the loss on the answer only. Training alternates between the block at the start and at a random position; the two test sets are identical except for that position. With the block at the start, SSA must keep its 17 chunks through about 480 eviction decisions. We report token accuracy per answer token excluding the null symbol, where chance is about 25\%. Table~\ref{tab:haystack} reports each model after 1,760 updates on this task, started from its mixed-task checkpoint or from its $k$-hop checkpoint, and after 3,520 updates from the mixed-task checkpoint.

\begin{table}[h!]
\centering\small
\caption{$k$-hop induction with the source block hidden in an 8K input: answer-token accuracy excluding the null symbol (\%), overall and at 4 / 8 / 16 hops in parentheses. Chance is about 25\%. Models train for 1,760 updates from the named checkpoint, except in the last block.}
\label{tab:haystack}
\setlength{\tabcolsep}{3.5pt}
\begin{tabular}{@{}llcc@{}}
\toprule
Model & Starting checkpoint & Block at the start & Block at a random position \\
\midrule
SSA & mixed-task & 77.4 (100.0 / 99.9 / 25.9) & 26.3 (27.0 / 26.3 / 25.6) \\
Transformer $576\rightarrow576$ & mixed-task & 22.5 (21.3 / 22.0 / 24.6) & 22.7 (21.3 / 22.3 / 24.9) \\
Transformer $8\mathrm{K}\rightarrow8\mathrm{K}$ & mixed-task & 26.7 (27.9 / 26.3 / 25.8) & 26.6 (27.4 / 26.4 / 25.9) \\
GDN & mixed-task & 26.5 (26.7 / 26.9 / 26.0) & 26.4 (26.8 / 26.4 / 25.9) \\
HOLA & mixed-task & 26.5 (27.1 / 26.2 / 25.9) & 26.3 (27.0 / 26.0 / 25.8) \\
\midrule
SSA & $k$-hop (Table~\ref{tab:posttraining_final}) & 99.2 (99.4 / 99.3 / 98.9) & 26.7 (27.5 / 26.4 / 25.9) \\
Transformer $576\rightarrow576$ & $k$-hop (Table~\ref{tab:posttraining_final}) & 22.6 (21.1 / 22.3 / 24.6) & 22.9 (21.5 / 22.6 / 24.9) \\
Transformer $8\mathrm{K}\rightarrow8\mathrm{K}$ & $k$-hop (Table~\ref{tab:posttraining_final}) & 99.9 (100.0 / 100.0 / 99.8) & 26.8 (27.5 / 26.8 / 26.0) \\
GDN & $k$-hop (Table~\ref{tab:posttraining_final}) & 27.2 (28.2 / 27.1 / 26.2) & 26.9 (28.1 / 26.7 / 25.9) \\
HOLA & $k$-hop (Table~\ref{tab:posttraining_final}) & 27.0 (28.2 / 26.9 / 25.9) & 26.9 (27.8 / 26.7 / 25.9) \\
\midrule
SSA & mixed-task, 3,520 updates & 85.2 (86.4 / 85.7 / 83.3) & 26.8 (28.3 / 26.2 / 25.7) \\
Transformer $576\rightarrow576$ & mixed-task, 3,520 updates & 22.8 (21.6 / 22.6 / 24.6) & 22.7 (21.4 / 22.4 / 24.7) \\
Transformer $8\mathrm{K}\rightarrow8\mathrm{K}$ & mixed-task, 3,520 updates & 53.1 (98.6 / 29.7 / 25.7) & 26.8 (27.6 / 26.6 / 26.0) \\
GDN & mixed-task, 3,520 updates & 27.1 (28.2 / 26.9 / 25.9) & 27.0 (28.0 / 26.8 / 26.2) \\
HOLA & mixed-task, 3,520 updates & 27.3 (28.6 / 27.2 / 25.8) & 27.2 (28.3 / 27.1 / 25.9) \\
\bottomrule
\end{tabular}
\end{table}

With the block at the start, SSA from its $k$-hop checkpoint reaches 99.2\% against 99.9\% for the full-attention Transformer, and 98.9\% against 99.8\% at 16 hops, while Transformer-$576$, GDN and HOLA stay at chance from either checkpoint. From the mixed-task checkpoint SSA is the only model above chance after 1,760 updates (77.4\%, with 16 hops at chance); after 3,520 updates it reaches 85.2\% (83.3\% at 16 hops) and the 8K Transformer 53.1\% (98.6\% at 4 hops, 29.7\% and 25.7\% at 8 and 16). With the block at a random position every model is at chance in every setting.

A retention probe on SSA with the block at a random position (192 rows) finds that 97.1\% of the block's chunks are in memory at the first answer step, and that the whole block stays in memory at every answer step in 51\% of rows. Token accuracy excluding the null symbol is 26.4\% when the whole block was kept and 26.3\% when it was not. SSA's memory therefore keeps the block, and its reader fails to use a block whose offset varies. The full-attention Transformer fails in the same setting with the whole input readable.

\subsection{What the retention policy keeps during language modeling}
\label{app:lm_diagnostics}

\paragraph{Retention probe.}
We record the scorer's decisions on 128 windows of 8,192 tokens from the held-out SlimPajama corpus. The pretrained SSA without recovery applies a fixed positional rule, identical in all 128 windows. Chunk 0 is resident at every step; at the last step 15 of the 34 residents lie more than 4K tokens back and the other 19 are the 19 most recent chunks. The post-trained SSA (trainable scorer, Table~\ref{tab:posttraining_final}) is selective and varies with the text. Its overlap with a sliding window over the 34 most recent chunks averages 0.23 (0.08 to 0.77 across windows), it evicts the incoming candidate in 66\% of decisions, and at the last step its residents are spread over the whole input: 16\%, 11\%, 22\%, 25\% and 26\% lie 0 to 576, 576 to 1K, 1K to 2K, 2K to 4K and 4K to 8K tokens back.

\paragraph{Forced retention rules.}
Table~\ref{tab:forced_window} keeps the weights and replaces the learned rule, at evaluation only, with one of two fixed rules: a sliding window over the 34 most recent chunks, or the first chunk plus the 33 most recent. Every key keeps its original position.

For the pretrained model, the sliding window removes the rise from 2K to 8K (14.94 at 2K and 14.83 at 8K, against 14.23 and 76.24 with the learned rule). Swapping one recent chunk for the first chunk brings perplexity at 8K back up to 42.06. That chunk is read from up to 8K tokens back, far beyond the distances seen in 2K pretraining, so these long distances are the likely cause of the rise. At 2K, the learned rule is slightly better than the sliding window (14.23 against 14.94).

For the post-trained model, the sliding window lowers perplexity from 23.08 to 20.81 at 2K and from 27.08 to 20.16 at 8K, so keeping old chunks costs about 2.3 points at 2K and 7 at 8K.

These fixed rules are diagnostics only. A sliding window gives up the long-range retrieval that the learned rule is trained for; App.~\ref{app:window_size} tests the middle ground, a wider local window next to a learned memory.

\begin{table}[h!]
\centering\small
\caption{Held-out perplexity of SSA with its learned retention rule and with two fixed rules forced at evaluation. The pretrained learned values are those of Table~\ref{tab:scaling} (2.656 and 4.334 nats). The post-trained values come from a different GPU type than Table~\ref{tab:posttraining_final}, changing the learned-rule NLL by under 0.2\%.}
\label{tab:forced_window}
\begin{tabular}{@{}llrr@{}}
\toprule
Model & Retention rule at evaluation & PPL at 2K & PPL at 8K \\
\midrule
Pretrained SSA, no recovery & learned & 14.23 & 76.24 \\
 & sliding window, 34 most recent chunks & 14.94 & 14.83 \\
 & first chunk and 33 most recent & 14.14 & 42.06 \\
\midrule
Post-trained SSA, trainable scorer & learned & 23.08 & 27.08 \\
 & sliding window, 34 most recent chunks & 20.81 & 20.16 \\
 & first chunk and 33 most recent & 20.58 & 20.87 \\
\bottomrule
\end{tabular}
\end{table}

\paragraph{Capping query--key distances.}
At 8K, retained chunks are read at distances beyond those seen in 2K pretraining. Table~\ref{tab:distance_cap} caps them at evaluation only: a key more than 2,032 tokens before the query chunk is read as if it were 2,032 tokens before, so no distance exceeds 2,047. The cap removes about half of SSA's rise from 2K to 8K, and all of the 2K-pretrained Transformer's.

\begin{table}[h!]
\centering\small
\caption{Held-out perplexity with query--key distances capped at the pretraining length, at evaluation only.}
\label{tab:distance_cap}
\begin{tabular}{@{}lrrr@{}}
\toprule
Model & 2K & 8K & 8K, distances capped \\
\midrule
SSA, trainable scorer & 23.07 & 26.91 & 24.92 \\
Transformer $2\mathrm{K}\rightarrow8\mathrm{K}$ & 19.73 & 22.28 & 18.77 \\
\bottomrule
\end{tabular}
\end{table}

\paragraph{Open-ended generation.}
In a small check of open-ended generation, every 400M model still writes fluent English after post-training, but post-trained models sometimes fall into repetition loops when sampling, and after the $k$-hop stage every model, whatever its architecture, sometimes writes strings of the $k$-hop letters a--d (for example, to complete a URL); as expected, no pretrained model does either. For specific details, see the code.

\subsection{Local window size and language modeling}
\label{app:window_size}

SSA's language modeling cost (Table~\ref{tab:forced_window}) raises the question of how much recent context a bounded reader needs. We probe this with the post-trained SSA (trainable scorer, Table~\ref{tab:posttraining_final}), at evaluation only and on the same held-out corpus. Each query chunk reads $R$ chunks: itself, the $w$ chunks before it, and $M=R-w-1$ older chunks. At SSA's budget ($R=36$) we move chunks from memory to the window in three ways. With learned memory, the $w-1$ lowest-scored of SSA's 34 memory chunks are dropped and the $w-1$ most recent chunks before its one-chunk window are read instead, so memory keeps the scorer's $M$ highest-scored choices. To separate the window from what the scorer keeps, we also bypass the scorer and fill memory by one of two content-blind rules: random eviction, under which memory holds mostly recent chunks, or a reservoir sample, a uniform sample of every chunk that has left the window, under which memory holds mostly old chunks. Keys keep their original positions, and we also report perplexity with query--key distances capped at 2,047 tokens as in Table~\ref{tab:distance_cap}.

\begin{table}[h!]
\centering\small
\caption{Held-out perplexity at 8K with SSA's 576-token budget ($R=36$ chunks), by window size $w$, with memory kept by SSA's scorer, by random eviction or by a reservoir sample. The last row reads no memory: a sliding window over the 35 most recent chunks. The learned-memory row at $w=1$ is SSA's original evaluation (Tables~\ref{tab:posttraining_final} and~\ref{tab:distance_cap}); the other learned-memory rows come from a probe on a different GPU type, on which the $w=1$ values are 27.08 and 25.06.}
\label{tab:window_size}
\setlength{\tabcolsep}{4.5pt}
\begin{tabular}{@{}rrrrrrrr@{}}
\toprule
& & \multicolumn{2}{c}{Learned memory} & \multicolumn{2}{c}{Random memory} & \multicolumn{2}{c}{Reservoir memory} \\
\cmidrule(lr){3-4}\cmidrule(lr){5-6}\cmidrule(l){7-8}
$w$ & $M$ & Uncapped & Capped & Uncapped & Capped & Uncapped & Capped \\
\midrule
1 & 34 & 26.91 & 24.92 & 20.73 & 20.58 & 29.39 & 26.43 \\
5 & 30 & 23.40 & 21.91 & 20.40 & 20.32 & 24.41 & 22.22 \\
9 & 26 & 22.26 & 21.00 & 20.29 & 20.25 & 23.13 & 21.18 \\
13 & 22 & 21.68 & 20.59 & 20.19 & 20.18 & 22.57 & 20.77 \\
17 & 18 & 21.29 & 20.38 & 20.16 & 20.16 & 22.13 & 20.49 \\
25 & 10 & 20.81 & 20.19 & 20.14 & 20.14 & 21.56 & 20.22 \\
\midrule
35 & 0 & 20.17 & 20.17 & 20.17 & 20.17 & 20.17 & 20.17 \\
\bottomrule
\end{tabular}
\end{table}

When memory holds old content, the reader needs a local window of about 13 chunks (208 tokens) for good language modeling, and a wider window changes perplexity only a little (Table~\ref{tab:window_size}), as earlier truncation studies found for LSTM and Transformer language models \citep{khandelwal2018sharp,press2021shortformer}. With SSA's learned memory and distances capped, widening the window from 1 to 13 chunks lowers perplexity from 24.92 to 20.59, and no wider window lowers it by more than 0.44; at 2K it falls from 23.07 to 20.95, and by at most 0.15 more. Reservoir memory behaves the same way (26.43 at $w=1$, 20.77 at $w=13$, 0.55 lower at $w=25$). With random memory, which already holds mostly recent chunks, the window matters less, and from $w=13$ on perplexity matches the sliding window to within 0.03. Without the cap, perplexity with learned memory keeps falling past $w=13$ (21.68 at $w=13$ against 20.17 with no memory); the cap removes most of that difference, which points to old chunks read at distances beyond the 2K pretraining length.

Across budgets of $R=36$, 72, 144 and 288 chunks (reservoir memory, distances capped), going from $w=1$ to $w=13$ lowers perplexity by 5.66, 3.58, 1.82 and 0.40, and widening the window further, to the largest $w$ each budget allows, lowers it by at most 0.55. The window also keeps language modeling stable with context length: at the 576-token budget with reservoir memory and distances capped, perplexity at 2K, 8K, 16K and 32K is 22.63, 26.43, 28.50 and 30.54 with $w=1$, against 20.93, 20.77, 20.94 and 21.04 with $w=13$.

Our 400M models use $w=1$. With learned memory, $w=13$ recovers most of the gap between $w=1$ and a pure sliding window at the same budget: 2.12 of 2.23 perplexity points at 2K, and 4.33 of 4.75 at 8K with distances capped. These probes measure language modeling only, and we have not tested them on the benchmarks, as many benchmarks were made to barely fit within SSA's context.

\section{Training and inference cost}
\label{app:compute}

\paragraph{Training.}
Table~\ref{tab:cost_runs} gives the time per update of our 400M training runs. In pretraining, the hard-phase value with recovery is from the recovery-pretrained model of Table~\ref{tab:scaling}, averaged over 12,714 hard-phase updates on the same hardware; its soft phase took 10.4\,s per update. Recovery more than doubles a hard update there. In the post-training pilot, recovery selects 8 of each hard update's 64 rows and replays them in one pass every 8 hard updates, which adds 20.7 seconds per hard update (45\%). In post-training, the soft phase is slower than the hard phase without recovery, likely because it also computes the attention-distillation loss and reads every chunk. The post-training Transformer value is from the 2K-pretrained full-attention Transformer on the same task mixture and GPU type.

\begin{table}[h!]
\centering\small
\caption{Time per update in the 400M training runs, with the ratio to the Transformer in parentheses. GPU types differ between rows.}
\label{tab:cost_runs}
\begin{tabular}{@{}p{0.33\linewidth}rrrr@{}}
\toprule
& & \multicolumn{3}{c}{SSA} \\
\cmidrule(l){3-5}
Setting & Transformer & Soft phase & Hard, no recovery & Hard, recovery \\
\midrule
Pretraining at 2K, 4 H100, 524,288 tokens per update & 2.5\,s & 9.1\,s (3.6$\times$) & 14.9\,s (6.0$\times$) & 32\,s (13$\times$) \\
Post-training pilot, mixed tasks up to 8K, 2 RTX PRO 6000, 64 rows per update & 19.9\,s & 59.5\,s (3.0$\times$) & 45.7\,s (2.3$\times$) & 66.4\,s (3.3$\times$) \\
\bottomrule
\end{tabular}
\end{table}

\paragraph{Cost grid: protocol.}
We time SSA against a full-attention Transformer at six model shapes and six context lengths, from 2K to 64K tokens, with random weights and random token ids. The shapes (layers, width, heads) are (4, 256, 4), (8, 512, 8), (12, 768, 12), (16, 1280, 20), (24, 1024, 16) and (22, 2048, 32), with 3M, 25M, 85M, 315M, 367M and 1.11B parameters outside the embedding table (32,000-token vocabulary); every head has dimension 64. Our 400M models have the fifth shape with a different head layout, which we report separately below. At every size SSA reads 34 chunks of 16 tokens, one window chunk and the current chunk (576 tokens), its scorer has two layers and 3.45\% of the parameters, and its reader's feed-forward width is reduced until SSA's total matches the Transformer's, as in the trained 400M pair. All cells run on one NVIDIA RTX PRO 6000 GPU with 96\,GB.

A training cell is one complete update (forward, backward, gradient clipping and AdamW) on one sequence, with a next-token loss at every position, BF16 autocast, gradient checkpointing and reader dropout 0.1, without recovery. The soft phase is measured mid-anneal ($\lambda=0.5$, $\tau=\sqrt{0.05}$), where every earlier chunk is read. We time three updates after two warm-up updates, and peak memory covers the timed updates. The scorer is re-scored with gradients over its recorded decisions 32 at a time. This gives the same gradients as one pass over all decisions and keeps the scorer's activation memory from growing with the number of decisions.

An inference cell reads a prompt of the context length minus 256 tokens and then generates 256 tokens greedily at batch size 1. Both models use optimized decoders (fixed key--value buffers, BF16 weights, CUDA graphs and compiled layers), and SSA reads the prompt 32 chunk steps at a time. We report medians over three prompts after one warm-up prompt.

\begin{table}[h!]
\centering\small
\caption{Cost of SSA relative to a full-attention Transformer of the same shape and parameter count (SSA / Transformer; below 1 means SSA is cheaper), by model size (parameters outside the embedding table) and context length. Every shape uses attention heads of dimension 64. SSA reads 576 tokens at every context length. Training: time per update in the hard phase, without recovery. Decoding: time per generated token after the prompt. One GPU, protocol and further tables in App.~\ref{app:compute}.}
\label{tab:cost_grid}
\setlength{\tabcolsep}{3.5pt}
\begin{tabular}{@{}lrrrrrrrrrrrr@{}}
\toprule
& \multicolumn{6}{c}{Training time per update} & \multicolumn{6}{c}{Decoding time per token} \\
\cmidrule(lr){2-7}\cmidrule(l){8-13}
Parameters & 2K & 4K & 8K & 16K & 32K & 64K & 2K & 4K & 8K & 16K & 32K & 64K \\
\midrule
3M & 12.8 & 18.5 & 24.2 & 25.6 & 21.5 & 16.6 & 1.29 & 1.26 & 1.17 & 1.10 & 0.91 & 0.56 \\
25M & 6.7 & 11.7 & 14.5 & 13.7 & 10.3 & 7.1 & 1.17 & 1.10 & 0.98 & 0.72 & 0.54 & 0.36 \\
85M & 5.7 & 8.3 & 10.1 & 9.1 & 6.7 & 4.7 & 1.16 & 0.99 & 0.86 & 0.67 & 0.47 & 0.29 \\
315M & 5.2 & 6.6 & 7.5 & 6.5 & 5.0 & 3.5 & 0.99 & 0.90 & 0.76 & 0.58 & 0.39 & 0.24 \\
367M & 5.0 & 6.5 & 6.8 & 6.0 & 4.6 & 3.2 & 0.97 & 0.89 & 0.77 & 0.59 & 0.40 & 0.25 \\
1.11B & 4.4 & 5.7 & 6.2 & 5.5 & 4.5 & 3.1 & 0.96 & 0.87 & 0.74 & 0.57 & 0.39 & 0.24 \\
\bottomrule
\end{tabular}
\end{table}

\paragraph{Cost grid: training.}
SSA is slower per update in every cell (Table~\ref{tab:cost_grid} for the hard phase, Table~\ref{tab:cost_soft_prompt} for the soft phase). In the hard phase SSA's time per token is flat in context, since its read is bounded, while the Transformer's grows, so the ratio falls with context beyond 8K at every size from 25M up. The soft phase reads every earlier chunk, and its ratio rises with context.

Hard-phase peak memory is within 17\% of the Transformer's at the 3M, 25M, 85M and 367M shapes (Table~\ref{tab:cost_memory}). It reaches 1.36 times at 315M and 1.57 times at 1.11B. The soft phase holds dense attention maps over all earlier chunks, one per head, so its extra memory grows with the number of heads and the square of the context length, not with model size: at 32K it is 7.5 to 7.9\,GiB per head in the configurations that fit. The 4-head shape (3M) peaks at 44\,GiB there and the 8-head shape (25M) at 77\,GiB, and the shapes with 12 or more heads run out of memory on the 96\,GB GPU; at 64K every shape does.

\begin{figure}[h!]
    \centering
    \includegraphics[width=0.85\linewidth]{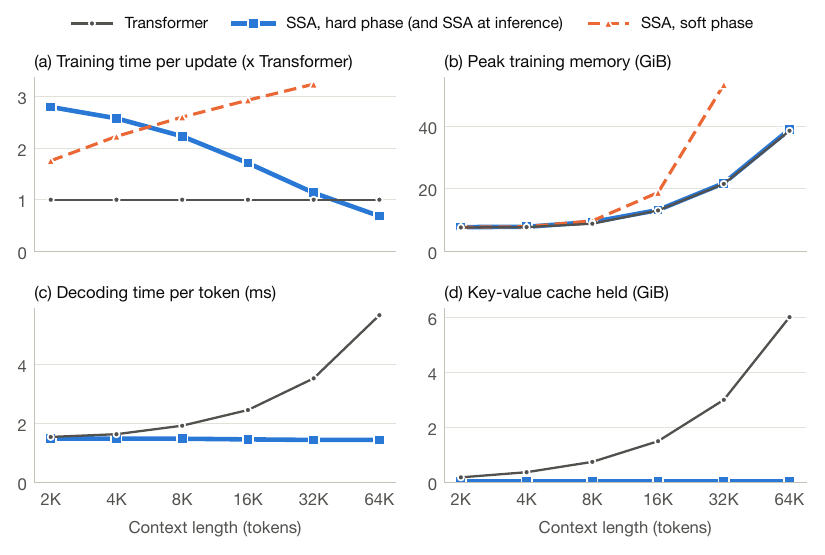}
    \caption{SSA against a full-attention Transformer for our 400M configuration (4 heads of dimension 256), by context length. (a) Training time per update relative to the Transformer. (b) Peak training memory. (c) Decoding time per generated token. (d) Key--value cache held at inference. The soft phase runs out of memory at 64K.}
    \label{fig:cost}
\end{figure}

\paragraph{Our 400M configuration.}
Our 400M models share the 367M shape but use 4 attention heads of dimension 256, as do the GDN and HOLA models we compare with. All shapes run the same code. On this layout the Transformer's training update is 1.7 times slower at 2K and 4.0 times slower at 64K than with 16 heads of dimension 64, while SSA's is 1.1 to 1.2 times faster, so the ratios for this configuration are lower than those of Table~\ref{tab:cost_grid}: the hard-phase ratio falls from 2.8 at 2K to 1.1 at 32K and 0.68 at 64K (12.0 against 17.6\,s per update; Table~\ref{tab:cost_train_400m}, Figure~\ref{fig:cost}). On this layout the hard phase without recovery is thus faster than the Transformer at 64K; we did not time recovery at that length. The training times of Table~\ref{tab:cost_runs} are for this configuration. Decoding ratios do not depend on the head layout (Table~\ref{tab:inference_scaling}). Its hard-phase peak memory is within 6\% of the Transformer's at every context length (Table~\ref{tab:cost_train_400m}).

\begin{table}[h!]
\centering\small
\caption{Training cost of our 400M configuration (the 367M shape with 4 heads of dimension 256): time per update on one sequence, with the ratio to the Transformer in parentheses, and peak allocated memory. ``oom'': the cell ran out of memory.}
\label{tab:cost_train_400m}
\begin{tabular}{@{}lrrrrrr@{}}
\toprule
& \multicolumn{3}{c}{Time per update (s)} & \multicolumn{3}{c}{Peak memory (GiB)} \\
\cmidrule(lr){2-4}\cmidrule(l){5-7}
Context & Transformer & SSA, hard & SSA, soft & Transformer & SSA, hard & SSA, soft \\
\midrule
2K & 0.14 & 0.39 (2.8$\times$) & 0.25 (1.8$\times$) & 7.6 & 7.7 & 7.7 \\
4K & 0.28 & 0.72 (2.6$\times$) & 0.62 (2.2$\times$) & 7.7 & 7.8 & 7.8 \\
8K & 0.63 & 1.40 (2.2$\times$) & 1.64 (2.6$\times$) & 8.8 & 9.3 & 9.7 \\
16K & 1.67 & 2.87 (1.7$\times$) & 4.91 (2.9$\times$) & 13.1 & 13.2 & 18.8 \\
32K & 5.19 & 5.89 (1.1$\times$) & 16.85 (3.2$\times$) & 21.6 & 21.8 & 53.3 \\
64K & 17.65 & 12.03 (0.7$\times$) & oom & 38.6 & 39.0 & oom \\
\bottomrule
\end{tabular}
\end{table}

\begin{table}[h!]
\centering\small
\caption{SSA / Transformer ratios by model size and context length: soft-phase training time per update, and time to read the prompt at inference. ``oom'': the SSA cell ran out of memory.}
\label{tab:cost_soft_prompt}
\setlength{\tabcolsep}{3.5pt}
\begin{tabular}{@{}lrrrrrrrrrrrr@{}}
\toprule
& \multicolumn{6}{c}{Soft-phase training time} & \multicolumn{6}{c}{Prompt-reading time} \\
\cmidrule(lr){2-7}\cmidrule(l){8-13}
Parameters & 2K & 4K & 8K & 16K & 32K & 64K & 2K & 4K & 8K & 16K & 32K & 64K \\
\midrule
3M & 12.9 & 16.7 & 23.1 & 28.1 & 31.0 & oom & 11.0 & 34.1 & 38.8 & 41.5 & 33.2 & 21.3 \\
25M & 5.2 & 10.5 & 15.9 & 21.2 & 25.9 & oom & 10.7 & 12.1 & 17.6 & 14.9 & 11.3 & 7.0 \\
85M & 3.9 & 7.4 & 12.3 & 18.1 & oom & oom & 5.8 & 9.3 & 9.8 & 8.4 & 5.7 & 3.9 \\
315M & 3.4 & 5.7 & 10.0 & 14.7 & oom & oom & 5.0 & 5.8 & 6.3 & 4.9 & 3.6 & 2.6 \\
367M & 3.1 & 5.5 & 9.1 & 14.4 & oom & oom & 5.0 & 5.6 & 5.4 & 4.5 & 3.2 & 2.2 \\
1.11B & 2.9 & 5.0 & 8.4 & 12.9 & oom & oom & 4.1 & 4.7 & 4.6 & 3.7 & 2.9 & 2.1 \\
\bottomrule
\end{tabular}
\end{table}

\begin{table}[h!]
\centering\small
\caption{SSA / Transformer ratio of peak training memory by model size and context length. ``oom'': the SSA cell ran out of memory.}
\label{tab:cost_memory}
\setlength{\tabcolsep}{3.5pt}
\begin{tabular}{@{}lrrrrrrrrrrrr@{}}
\toprule
& \multicolumn{6}{c}{Hard phase} & \multicolumn{6}{c}{Soft phase} \\
\cmidrule(lr){2-7}\cmidrule(l){8-13}
Parameters & 2K & 4K & 8K & 16K & 32K & 64K & 2K & 4K & 8K & 16K & 32K & 64K \\
\midrule
3M & 1.00 & 1.00 & 1.00 & 1.00 & 1.00 & 1.00 & 1.02 & 1.04 & 1.09 & 1.66 & 3.12 & oom \\
25M & 1.00 & 1.00 & 1.00 & 1.01 & 1.01 & 1.01 & 1.02 & 1.04 & 1.47 & 2.69 & 5.20 & oom \\
85M & 1.12 & 1.06 & 1.01 & 1.01 & 1.01 & 1.01 & 1.09 & 1.19 & 1.82 & 3.41 & oom & oom \\
315M & 1.01 & 1.15 & 1.29 & 1.30 & 1.36 & 1.17 & 1.01 & 1.24 & 2.09 & 4.06 & oom & oom \\
367M & 1.01 & 1.06 & 1.17 & 1.14 & 1.17 & 1.01 & 1.01 & 1.13 & 1.78 & 3.28 & oom & oom \\
1.11B & 1.00 & 1.00 & 1.14 & 1.45 & 1.57 & 1.45 & 1.00 & 1.01 & 1.59 & 3.74 & oom & oom \\
\bottomrule
\end{tabular}
\end{table}

\paragraph{Cost grid: inference.}
SSA's decoding time per token is flat in context and the Transformer's grows, so the ratio in Table~\ref{tab:cost_grid} falls with context at every size. Table~\ref{tab:inference_scaling} gives our 400M configuration. SSA reads the prompt more slowly at every size (Table~\ref{tab:cost_soft_prompt}), because its eviction decisions are sequential. For our 400M configuration the extra prompt time is repaid by the faster decoding after about 380 generated tokens at 8K, 270 at 32K and 200 at 64K.

\begin{table}[h!]
\centering\small
\caption{Inference for our 400M configuration: decoding time per generated token, time to read the prompt, and key--value cache held.}
\label{tab:inference_scaling}
\begin{tabular}{@{}lrrrrrr@{}}
\toprule
& \multicolumn{2}{c}{Per token (ms)} & \multicolumn{2}{c}{Prompt (s)} & \multicolumn{2}{c}{Key--value cache (MiB)} \\
\cmidrule(lr){2-3}\cmidrule(lr){4-5}\cmidrule(l){6-7}
Context & Transformer & SSA & Transformer & SSA & Transformer & SSA \\
\midrule
2K & 1.54 & 1.49 & 0.011 & 0.046 & 192 & 54 \\
4K & 1.64 & 1.49 & 0.022 & 0.103 & 384 & 54 \\
8K & 1.92 & 1.48 & 0.046 & 0.214 & 768 & 54 \\
16K & 2.45 & 1.46 & 0.115 & 0.439 & 1536 & 54 \\
32K & 3.53 & 1.44 & 0.326 & 0.887 & 3072 & 54 \\
64K & 5.66 & 1.44 & 0.944 & 1.788 & 6144 & 54 \\
\bottomrule
\end{tabular}
\end{table}

\section{Multi-key variable tracking}
\label{app:mkvt}

\subsection{Data, objectives, and training schedules}

MKVT inserts statements such as \texttt{VAR abc = 41} into TinyStories prose and
queries them using \texttt{?abc=}. Keys are three lowercase letters drawn from the full
$26^3$ pool; keys beginning with \texttt{a} are eligible for querying. Values are distinct
two-digit strings within an example. We use four configurations
$(K,B)\in\{(4,16),(4,24),(8,8),(16,16)\}$, where $K$ is the number of eligible keys and
$B$ the total number of bindings. Each recipe is run on seeds 0, 1 and 2 of each configuration, giving twelve task--seed cases. The 1024-character target includes its statements and
answer suffix. A queried statement must occur more than seventeen pre-insertion document
characters before the document end; other eligible statements need not satisfy this
constraint. Thus statement placement is not fully independent of query eligibility. We impose this so that the task requires retention.

Models use the 97-character vocabulary of our earlier experiments. Training and validation stories are
disjoint, but adjacent procedural seeds generate overlapping example streams. The fixed final-task
validation set has 1024 examples per run. Whole-answer exact match requires both answer
characters to be correct under teacher forcing; the separate generation check uses greedy
decoding on 128 examples. The training objective adds separately averaged answer and
non-answer cross-entropies with unit weights, preventing the short answer from being
diluted by the much longer prose context.

\begin{table}[h!]
\centering\small
\caption{Completed Transformer MKVT experiments. Solved cases have teacher-forced
whole-answer exact match $\geq99\%$ on 1024 fixed-target validation examples.
Name randomization applies only during language training. The auxiliary source head
uses task annotations and is reported as a separate supervised control.}
\label{tab:mkvt_pretrain}
\begin{tabular}{@{}rllccc@{}}
\toprule
Language & Retrieval & Paired & Randomized & Source & Solved \\
characters & attention & queries & names & head & cases \\
\midrule
73M & NormSoftmax & Yes & No & No & 6/12 \\
73M & NormSoftmax & Yes & Yes & No & 3/12 \\
73M & NormSoftmax & Yes & No & Yes & 12/12 \\
998M & NormSoftmax & Yes & No & No & 12/12 \\
998M & Standard & Yes & No & No & 12/12 \\
998M & Standard & No & No & No & 10/12 \\
998M & NormSoftmax & Yes & Yes & No & 12/12 \\
998M & Standard & Yes & Yes & No & 12/12 \\
998M & Standard & No & Yes & No & 12/12 \\
\bottomrule
\end{tabular}
\end{table}

\begin{table}[h!]
\centering\small
\caption{MKVT language and retrieval budgets. The longer language phase adds training
before the same retrieval budget.}
\label{tab:mkvt_schedule}
\begin{tabular}{@{}lrr@{}}
\toprule
 & Short language phase & Extended language phase \\
\midrule
Epochs at each of lengths 256, 512, 1024 & 5 & 68 \\
Language examples per epoch & 8192 & 8192 \\
Language character presentations & 73,400,320 & 998,244,352 \\
Retrieval epochs & 135 & 135 \\
Total epochs & 150 & 339 \\
\bottomrule
\end{tabular}
\end{table}

Retrieval proceeds through $(L,K,B)=(128,2,2),(256,4,4),(512,\min(K,8),8)$ and then
the final task. The $(4,24)$ task adds an intermediate $(1024,4,16)$ stage. The first
retrieval stage is capped at 28 epochs, or 21 for $(4,24)$, and at least 51 epochs are
reserved for the final stage. Earlier stage transitions allocate the saved epochs to
the final task; total retrieval budgets are equal even though allocations across
stages can differ.

The Transformer has 19,028,992 parameters and SSA has 22,499,712, so they are backbone matched rather than full-parameter matched. Transformer runs use batch size 64, dropout 0.1, AdamW, gradient
clipping at one, and a cosine learning rate from $3\cdot10^{-4}$ to $10^{-5}$. The extended
language phase adds 189 epochs at the initial learning rate so that the retrieval schedule
is preserved. Paired batches contain 32 contexts with two eligible queries each; unpaired
batches contain 64 independent contexts. All evaluations are unpaired. NormSoftmax,
when used, is restricted to retrieval training. Randomized-name pretraining replaces
repeated story names consistently within each document; retrieval prose is unmodified.

SSA uses unpaired queries, dropout 0.1, constant learning rate
$10^{-4}$, and recovery during retrieval. The extended schedule places 136 soft-retention
epochs within language training, followed by hard retention; retrieval begins at epoch
205. The four arms cross name randomization with recovery during the language phase.
The longer hard-language interval also reduces the sampling temperature at retrieval
onset relative to the shorter schedule.

\subsection{Retrieval onset and diagnostic controls}

\paragraph{Retrieval learning can fail without memory loss.}
An ablation study of 24 Transformer recipes found substantial seed sensitivity at small scale: the
best shared recipe solved six of twelve cases, and no recipe solved all three seeds of
any task. Failed checkpoints often produced a candidate value without responding to
which key was queried. In a paired diagnostic, one failed checkpoint answered only
eight of 64 queries correctly across eight shared passages, whereas a successful
checkpoint answered all 64. The failed model returned the same answer for different
queries about each passage. Since the Transformer retains the full context, this
failure cannot be explained by eviction. It indicates a failure to learn query-dependent
selection, consistent with the delayed emergence of attention operations observed in
other multi-step tasks \citep{hoffmann2024eureka}.

\paragraph{Language pretraining improves retrieval learning.}
We extend the language phase while preserving the retrieval budget and its learning-rate
schedule. A second intervention randomizes repeated proper names consistently within
each language-training story, which is intended to teach the model to respond to these randomizations. It changes the language examples, without using binding
locations or other task specific annotations. The retrieval syntax and evaluation tasks remain
unchanged.

Extended pretraining solves all development cases with paired queries under either
standard attention or NormSoftmax \citep{hoffmann2024eureka}, which we tried as a remedy for this failure (Table~\ref{tab:mkvt_pretrain}). Adding randomized
names also solves all cases with standard attention and unpaired queries. Most of the improvement comes from the extended language budget; randomized names reduce the number of solved cases at the short budget (3/12 against 6/12) and add two cases with standard attention and unpaired queries at the extended budget (12/12 against 10/12).

Table~\ref{tab:onset} reports first-stage retrieval onset for three extended-pretraining
variants with randomized names, each with twelve cases. The short-pretraining name
control changes names without increasing the language budget and solves $(4,16)$ seed 1
and $(8,8)$ seeds 0 and 2.

\begin{table}[h!]
\centering\small
\caption{Retrieval onset with extended language pretraining and randomized names.
Onset means first-stage teacher-forced answer exact match $\geq90\%$. Final success
uses the separate fixed-target metric.}
\label{tab:onset}
\begin{tabular}{@{}lrrr@{}}
\toprule
Retrieval configuration & By epoch 2 & Before stage exit & Final solved \\
\midrule
NormSoftmax, paired & 11/12 & 12/12 & 12/12 \\
Standard attention, paired & 11/12 & 12/12 & 12/12 \\
Standard attention, unpaired & 10/12 & 12/12 & 12/12 \\
\midrule
All three & 32/36 & 36/36 & 36/36 \\
\bottomrule
\end{tabular}
\end{table}

\paragraph{Task learning and language retention are distinct.}
As expected, improved retrieval and task performance does not guarantee preserved language modeling, and we observe a slight drop in language modeling performance. For the completed
Transformer arms with extended pretraining on unmodified stories, mean held-out prose
loss increases from about 0.370 nats per character at the end of language training to
0.500--0.512 after retrieval training. We therefore evaluate language loss separately
from retrieval accuracy.
\subsection{SSA results and source availability}

Table~\ref{tab:mkvt_ssa} reports the completed extended-pretraining suite at epoch 339.

\begin{table}[h!]
\centering\small
\caption{SSA MKVT at the end of training (epoch 339). Each arm contains twelve task--seed cases.
Exact is fixed-final-task teacher-forced whole-answer exact on 1024 validation examples.}
\label{tab:mkvt_ssa}
\begin{tabular}{@{}lrrr@{}}
\toprule
Language intervention & $\geq99\%$ & $\geq97.5\%$ & Lowest exact (\%) \\
\midrule
Extended pretraining & 11/12 & 12/12 & 98.0 \\
+ Randomized names & 9/12 & 12/12 & 97.6 \\
+ Recovery in language & 9/12 & 11/12 & 83.9 \\
+ Both & 11/12 & 11/12 & 86.1 \\
\midrule
All arms & 40/48 & 46/48 & 83.9 \\
\bottomrule
\end{tabular}
\end{table}

Forty-four runs reach 90\% hard first-stage answer exact within one to five retrieval
epochs.

Source availability is measured using recorded chunk memberships. A statement is directly
readable when any overlapping chunk remains in memory or the local window at the query.
A longer path through retained contextualized chunks denotes possible indirect access. We have not been able to establish a formal causal use relationship here.

A statement spans two or three chunks. Retaining the chunk with the value can be enough
to answer even when most statement characters are absent. A character-fraction metric
that excludes the local window can therefore understate usable retention. The chunk
availability diagnostic captures this distinction. In the earlier sample, cases with
one or two intermediate chunks comprise 2.9\% and 1.3\% of examples, with conditional
answer exact of 63.1\% and 46.8\%. Most successful retrieval is associated with direct
source availability.
\section{Additional related work}
\label{app:related}

\paragraph{Sparse reading and memory evaluation.} Quest and Native Sparse Attention reduce the amount of history read by a query, targeting the quadratic compute cost of attention \citep{tang2024quest,yuan2025nsa}. Selecting what to read and selecting what to retain are different constraints: restricting reads does not by itself bound the stored history. Copying, associative retrieval, and multi-hop induction provide controlled ways to examine these distinctions \citep{jelassi2024repeat,sanford2024transformers}. Our optimization analysis is also related to delayed learning of intermediate attention operations in multi-step tasks, which also appears in our training \citep{hoffmann2024eureka}. We use these tasks to distinguish reader learning from information loss caused by retention decisions. Cache selection and sparse reading address memory and compute separately; SSA, like SSMs, addresses both.

\end{document}